\documentclass[lettersize,journal]{IEEEtran}
\ifCLASSOPTIONcompsoc
  \usepackage[nocompress]{cite}
\else
  \usepackage{cite}
\fi
\ifCLASSINFOpdf
\else
\fi

\usepackage{times}
\usepackage{epsfig}
\usepackage{graphicx}
\usepackage{amsmath}
\usepackage{amssymb}
\usepackage{array}
\usepackage{makecell}

\usepackage{mathtools}
\usepackage{booktabs}
\usepackage{xcolor}
\usepackage{multirow}
\usepackage{float}
\usepackage{amsfonts}
\usepackage{comment}
\usepackage{url}
\usepackage{tabularx}
\usepackage{siunitx}
\usepackage[caption=false,font=normalsize,labelfont=sf,textfont=sf]{subfig}
\usepackage{diagbox} 

\usepackage[pagebackref=true,breaklinks=true,colorlinks,bookmarks=false]{hyperref}

\usepackage{algorithmic}
\usepackage{textcomp}
\usepackage{stfloats}
\usepackage{verbatim}
\def\BibTeX{{\rm B\kern-.05em{\sc i\kern-.025em b}\kern-.08em
    T\kern-.1667em\lower.7ex\hbox{E}\kern-.125emX}}
\usepackage{balance}

\usepackage[capitalize]{cleveref}
\crefname{section}{Sec.}{Secs.}
\Crefname{section}{Section}{Sections}
\Crefname{table}{Table}{Tables}
\crefname{table}{Tab.}{Tabs.}

\usepackage{pifont}
\newcommand{\etal}{\emph{et al.}} 
\newcommand{\li}{Li \emph{et al.}~\cite{cis2020}} 
\newcommand{\zhu}{Zhu \emph{et al.}~\cite{zhu2022montecarlo}} 

\begin{document}
\title{MAIR++: Improving Multi-view Attention Inverse Rendering with Implicit Lighting Representation}
\author{JunYong Choi,~\IEEEmembership{Student Member,~IEEE,}
        SeokYeong Lee,~\IEEEmembership{Student Member,~IEEE,}
        Haesol Park,~\IEEEmembership{Member,~IEEE,}
        Seung-Won Jung,~\IEEEmembership{Senior Member,~IEEE,}
        Ig-Jae Kim,~\IEEEmembership{Member,~IEEE,}
        Junghyun Cho,~\IEEEmembership{Member,~IEEE,}
\IEEEcompsocitemizethanks{\IEEEcompsocthanksitem All authors except Seung-Won Jung are with the Korea Institute of Science and Technology~(KIST), Seoul, Korea. Junyong Choi, SeokYeong Lee, Seung-Won Jung and Ig-Jae Kim are also with the Korea University. \protect\\
E-mail: \{happily, shapin94, haesol, drjay, jhcho\}@kist.re.kr, swjung83@korea.ac.kr \protect\\}}

\markboth{Journal of \LaTeX\ Class Files}%
{How to Use the IEEEtran \LaTeX \ Templates}

\maketitle

\begin{abstract}
In this paper, we propose a scene-level inverse rendering framework that uses multi-view images to decompose the scene into geometry, SVBRDF, and 3D spatially-varying lighting. While multi-view images have been widely used for object-level inverse rendering, scene-level inverse rendering has primarily been studied using single-view images due to the lack of a dataset containing high dynamic range multi-view images with ground-truth geometry, material, and spatially-varying lighting. To improve the quality of scene-level inverse rendering, a novel framework called Multi-view Attention Inverse Rendering (MAIR) was recently introduced. MAIR performs scene-level multi-view inverse rendering by expanding the OpenRooms dataset, designing efficient pipelines to handle multi-view images, and splitting spatially-varying lighting. Although MAIR showed impressive results, its lighting representation is fixed to spherical Gaussians, which limits its ability to render images realistically. Consequently, MAIR cannot be directly used in applications such as material editing. Moreover, its multi-view aggregation networks have difficulties extracting rich features because they only focus on the mean and variance between multi-view features. In this paper, we propose its extended version, called MAIR++. MAIR++ addresses the aforementioned limitations by introducing an implicit lighting representation that accurately captures the lighting conditions of an image while facilitating realistic rendering. Furthermore, we design a directional attention-based multi-view aggregation network to infer more intricate relationships between views. Experimental results show that MAIR++ not only achieves better performance than MAIR and single-view-based methods, but also displays robust performance on unseen real-world scenes.
\end{abstract}



\begin{figure*}[t!]
  \centering
  \includegraphics[width=\linewidth]{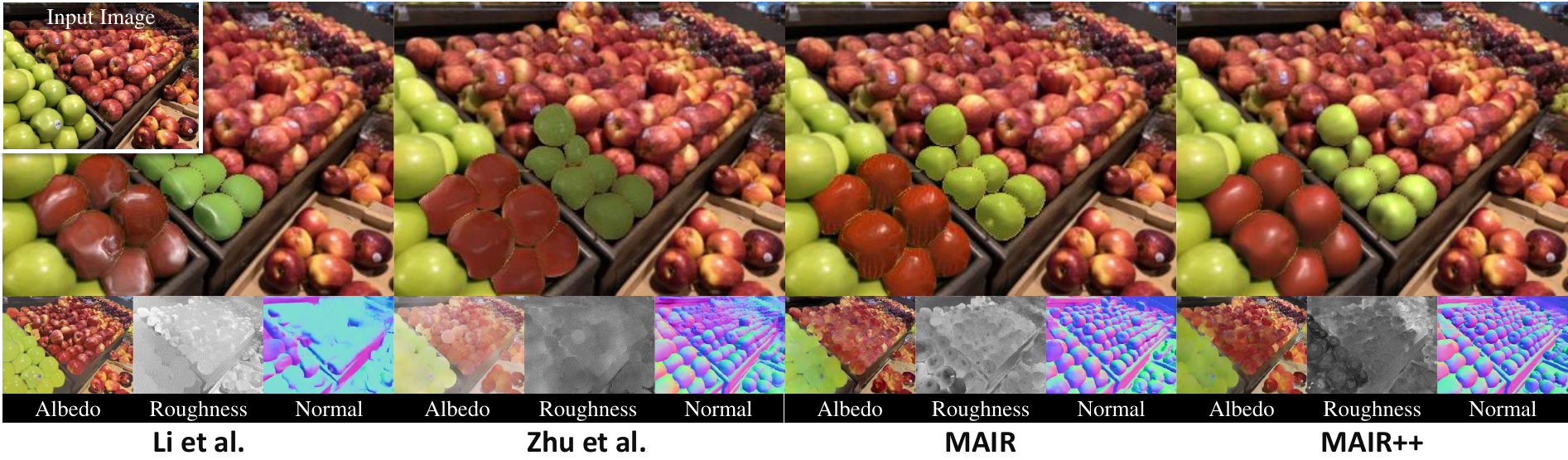}
  \caption{Experimental results on unseen real world data. First row: Material editing results that change the albedo of the apple. MAIR++ is the only method that preserves the apple's specularity and edits the material realistically. Second row: inverse rendering result. The single-view-based methods~\cite{cis2020, zhu2022montecarlo} rely solely on contextual information, making it difficult to estimate the complex materials and geometry of real-world scenes.}
   \label{fig:main}
\end{figure*}

\section{Introduction}\label{sec:introduction}
\IEEEPARstart{I}{nverse} rendering is a technology used to estimate material, lighting, and geometry from color images. Decomposing a scene through inverse rendering enables various applications, such as object insertion, relighting, and material editing in virtual reality (VR) and augmented reality (AR). However, since inverse rendering is an ill-posed problem, previous studies have focused only on a part of the inverse rendering, such as intrinsic image decomposition~\cite{id1,id2,id3,id4}, shape from shading~\cite{sfs1,sfs2,sfs3}, lighting estimation~\cite{gardner2019deep, li2023spatiotemporally, srinivasan2020lighthouse, weber2022editable, yu2023accidental, wang2022stylelight, zhan2021sparse}, and material estimation~\cite{sime1, sime2, sime3, sime4, sime5}. Conversely, we aim to perform full inverse rendering, including 3D spatially-varying lighting estimation.

Recent advances in GPU-accelerated physically-based rendering algorithms have enabled the construction of large-scale photorealistic indoor high dynamic range (HDR) image datasets that include geometries, materials, and spatially-varying lighting~\cite{openrooms2021, zhu2022montecarlo}. The availability of such datasets and the recent success of deep learning technology have led to seminal works on inverse rendering based on single view~\cite{cis2020, irisformer2022, vsg, phyir, Li22, zhu2022montecarlo}. However, these works have fundamental limitations in that they are prone to bias in the training dataset despite having shown promising results. Specifically, single-view-based inverse rendering estimates specular reflectance from the contextual information of the image, making it less reliable for predicting complex spatially-varying bidirectional reflectance distribution function (SVBRDF) in the real-world. Fig.~\ref{fig:main} shows such an example (left two columns), where decomposition has severely failed due to the complexity of the real-world scene. In addition, depth-scale ambiguity makes it challenging to employ these methods in 3D applications such as object insertion. 

\begin{figure*}[t!]
  \centering
  \includegraphics[width=\linewidth]{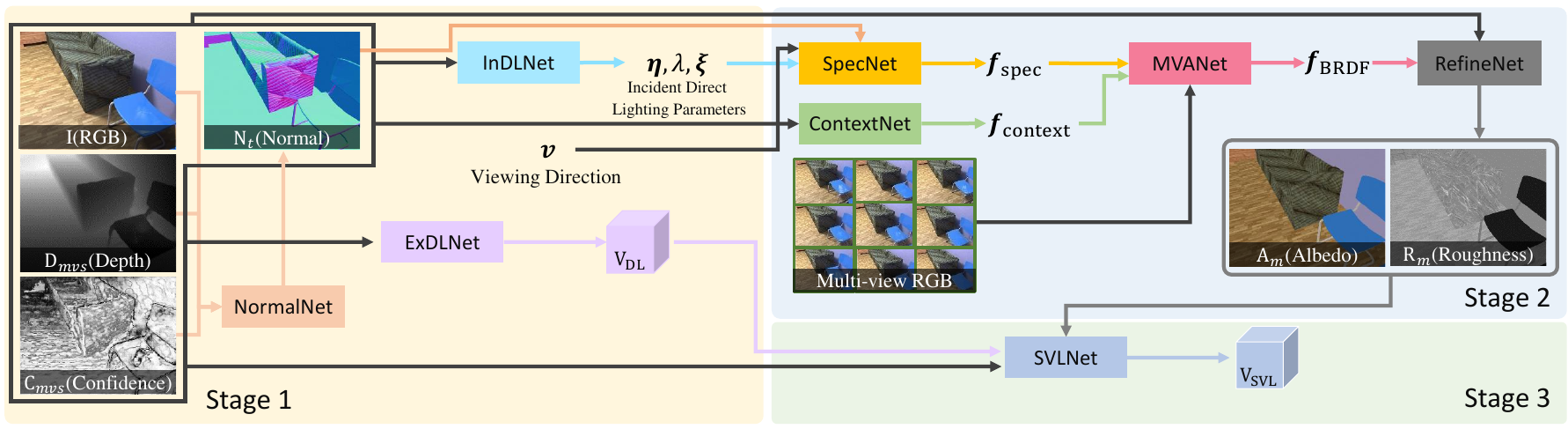}
  \caption{Entire pipeline of MAIR. MAIR addresses the difficulty of inverse rendering by splitting the scene components into the normal, direct lighting, material, and spatially-varying light and progressively estimating them.}
   \label{fig:mair_entire}
\end{figure*}

To address these problems, the \textbf{MAIR}~\cite{choi2023mair}, \textbf{M}ulti-view \textbf{A}ttention \textbf{I}nverse \textbf{R}endering framework, was recently introduced. MAIR exploits multiple RGB images of the same scene from multiple cameras and, more importantly, it utilizes multi-view stereo (MVS) depth as well as scene context to estimate SVBRDF. As a result, MAIR becomes less dependent on the implicit scene context and shows better performance on unseen real-world images. However, the processing of multi-view images inherently requires a high computational cost to handle multiple observations with occlusions and brightness mismatches. To remedy this, MAIR presents a three-stage training pipeline that can significantly increase training efficiency and reduce memory consumption. Spatially-varying lighting consists of direct lighting and indirect lighting. Indirect lighting affected by the surrounding environment makes inverse rendering difficult. Therefore, in Stage 1, MAIR first estimates the direct lighting and geometry, which reflect the amount of light entering each point and in which direction the specular reflection appears. MAIR then estimates the material in Stage 2 using the estimates of the direct lighting, geometry, and multi-view color images. In Stage 3, MAIR collects all the material, geometry, and direct lighting information and finally estimates 3D spatially-varying lighting, including indirect lighting. The pipeline of MAIR is shown in Fig.~\ref{fig:mair_entire}. MAIR also presents the OpenRooms Forward Facing (OpenRooms FF) dataset as an extension of OpenRooms~\cite{openrooms2021} to train the proposed network. 

Although MAIR can serve as a good starting point for scene-level multi-view inverse rendering, it has several limitations for higher-level inverse rendering tasks. First, MAIR mainly relied on direct lighting when inferring BRDF. This simplifies the inference of specular radiance, but limits its ability to accurately infer the BRDF. Additionally, MAIR's uniform sampling rendering based on spherical Gaussians fails to reproduce images accurately, making it less suitable for applications like material editing. Moreover, MAIR relies on MVS depth, which often has many holes and inaccuracies in complex indoor scenes. Lastly, MAIR's multi-view aggregation network focuses only on the mean and variance of multi-view features, which are insufficient for generating rich features needed to infer the BRDF.

This paper presents an extended version of MAIR, called MAIR++. The key differences in MAIR++ compared to MAIR can be summarized as follows. 

\begin{itemize}
\item We propose a novel lighting representation, Implicit Lighting Representation (ILR). ILR represents the full incident lighting of each pixel as a feature vector, which can be decoded as an environment map through a direction decoder, or can render images realistically through a neural renderer.

\item We train a single-view inverse rendering network to enhance the performance of MVS depth and to provide good initial values for multi-view inverse rendering.

\item We propose a novel Directional Attention Module (DAM) for multi-view aggregation network to thoroughly analyze ILR information from multi-view images. 

\item We propose the Albedo Fusion Module(AFM) for integrating single-view and multi-view albedo. This helps compensate for the shortcomings of the two albedo maps based on rendering.

\end{itemize}
Our ILR enables realistic material editing, which is challenging for MAIR. Fig.~\ref{fig:main} shows an example of this. MAIR's material editing exhibits artifacts due to uniform sampling rendering, does not accurately preserve existing specularity, and does not render the apple's shading realistically. In contrast, our method preserves the specularity of the apples and naturally adjusts the diffuse albedo. Additionally, our improved depth map, DAM, and AFM, along with ILR, enable more accurate inverse rendering.


The rest of this paper is organized as follows: In Section II, we introduce related work on MAIR++. Sections III and IV explain the structures of MAIR and MAIR++, respectively. Section V describes our OpenRooms FF dataset and training details. Section VI provides experimental results on inverse rendering, material editing, and object insertion and ablation studies, while Section VII discusses MAIR and MAIR++. Finally, Section VIII presents our conclusions about MAIR++.





\section{Related Works}
\noindent\textbf{Single-view inverse rendering.}
Research on inverse rendering has received significant attention in recent years owing to the development of deep learning technology. Yu \etal\cite{outdoorinv} performed outdoor inverse rendering with multi-view self-supervision, but their lighting is simple distant lighting. A pioneering work by \li\ conducted inverse rendering using a single image, and IRISformer\cite{irisformer2022} further improved the performance by replacing convolutional neural networks (CNNs) with a Transformer\cite{densevisiontransformer}. PhyIR \cite{phyir} addressed inverse rendering using a panoramic image. However, the lighting representation in~\cite{cis2020,irisformer2022, phyir} is a 2D per-pixel environment map, which is insufficient to model 3D lighting. Li \etal \cite{Li22} adopted a parametric 3D lighting representation; however, it was fixed with two types of indoor light sources. The recently introduced work by \zhu\ demonstrated realistic 3D volumetric lighting based on ray tracing. However, since these previous works\cite{outdoorinv, cis2020,irisformer2022,Li22, phyir, zhu2022montecarlo} are all based on a single view, they inherently rely on the context of the scene, making these methods less reliable for unseen images. In contrast, the proposed method can estimate BRDF and geometry more accurately by utilizing the multi-view correspondences as additional cues for inverse rendering. 
 
\noindent\textbf{Multi-view inverse rendering.}
Earlier works~\cite{maier2017intrinsic3d, zhang2016emptying} perform multi-view inverse rendering without deep learning, but require additional equipment to obtain RGB-D images. Philip \etal\cite{philip2019, philip2021} demonstrated a successful relighting with multi-view images; however, these methods cannot be used for applications such as object insertion because they use pretrained neural renderers. 3D geometry-based methods~\cite{cis-texture, invpath, kim2016multi} require additional computation to generate a mesh. PhotoScene~\cite{photoscene}, in particular, requires external CAD geometry, which is manually aligned for each object. With the advent of NeRF~\cite{nerf, mipnerf, mip360} and 3D Gaussian splatting (GS)~\cite{kerbl20233d}, many other optimization-based multi-view inverse rendering methods~\cite{wang2023neural, zhu2023i2, jin2023tensoir, refnerf, invrender2022, sun2023neural, liang2023gs, yao2022neilf, wu2023factorized} have also been proposed. Although all of these methods perform inverse rendering with high quality, they all require expensive time to obtain accurate fine geometry and radiance. In contrast, our method only needs per-view depth maps without requiring 3D geometry or test-time optimization, making it more computationally efficient and much easier to apply to more general scenes.

\noindent\textbf{Lighting estimation.}
Lighting estimation has been studied not only as a sub-task of the inverse rendering but also an important research topic~\cite{le1,le2,le3,Song19}. In Lighthouse \cite{lighthouse}, 3D spatially-varying lighting was obtained by estimating the RGBA lighting volume. Wang \etal\cite{vsg} proposed a more sophisticated 3D lighting volume, VSG(Volumetric Spherical Gaussians) by replacing RGB with a spherical Gaussian in the lighting volume. Because these methods~\cite{lighthouse,vsg} assume Lambertian reflectance, they cannot represent complex indirect lighting and have limitations in expressing HDR lighting due to weak-supervision with the LDR dataset\cite{interiornet}. On the other hand, the proposed method can handle complex SVBRDF and HDR lighting well because we train our model on the large indoor HDR dataset\cite{openrooms2021}. Li \etal\cite{li2023spatiotemporally} proposed a lighting representation that added uniform RGB to VSG and focused on spatiotemporally consistent lighting estimation for video. They were able to achieve high quality lighting, but this required a large number of high resolution HDR environment maps. On the other hand, a study on spatially-varying lighting estimation in outdoor scenes\cite{wang2022neural,SOLD-Net} was introduced. Because they focus on outdoor street scenes, they cannot clearly reproduce the indirect lighting by the scene material. 

\noindent\textbf{Neural rendering.}
With the development of neural rendering, papers related to inverse rendering or related applications are being proposed. NeRF-based papers that are robust to lighting changes~\cite{martin2021nerf, chen2022hallucinated, kuang2022neroic, lee2023extremenerf} have been proposed. They can remove the effects of lighting from an image, but have no control over it. ENVIDR~\cite{liang2023envidr} trained a neural renderer on the synthetic dataset and then used its prior to separate lighting from the material. However, this method target object-centric scenes and is not suitable for complex indoor scenes. Various results based on NeRF\cite{nerfeditable, nerfcomposite, nerfobjcen, giraffe, ye2023real}, such as scene editing or object insertion, have also been introduced. However, they did not consider the materials and lighting of the scene. In contrast, our work focuses on scene-level inverse rendering, which enables physically meaningful object insertion. Recently, inverse rendering papers~\cite{tewari2023diffusion, papantoniou2023relightify} based on the strong prior of the diffusion model have also been proposed. Lyu \etal~\cite{tewari2023diffusion} proposed an inverse rendering method based on diffusion posterior sampling. However, it is not suitable for object insertion because it does not take into account complex lighting such as spatially-varying lighting. Relightify~\cite{papantoniou2023relightify} was able to obtain the material by guiding the diffusion model with a known partial texture, but this did not take lighting into account. A paper using the diffusion model for object insertion~\cite{song2023objectstitch} also appeared. This shows reasonable insertion results based on a strong diffusion prior, but it is not stable and has clear limitations in its usability compared to inverse rendering methods.


\section{MAIR}\label{sec:method}
This section reviews our previous work, MAIR~\cite{choi2023mair}, which is a strong baseline for multi-view scene-level inverse rendering. Interested readers can see ~\cite{choi2023mair} for more details. 

Let $K$ be the number of viewpoints; then, the inputs to MAIR are $K$ triplets, where each triplet is composed of an RGB image $\mathrm{I} \in \mathbb{R}^{3 \times H \times W}$ with size H$\times$W, depth map $\mathrm{D}_{mvs}\in \mathbb{R}^{H \times W}$, and its confidence map $\mathrm{C}_{mvs} \in \mathbb{R}^{H \times W}$. $\mathrm{D}_{mvs}$ and $\mathrm{C}_{mvs}$ are obtained using a state-of-the-art MVS model\cite{giang2021curvature}. MAIR is embodied with a three-stage structure that progressively estimates the normal, direct lighting, material, and spatially-varying lighting. The entire pipeline of MAIR is presented in Fig.~\ref{fig:mair_entire}.

\begin{figure*}[ht]
  \centering
  \includegraphics[width=\linewidth]{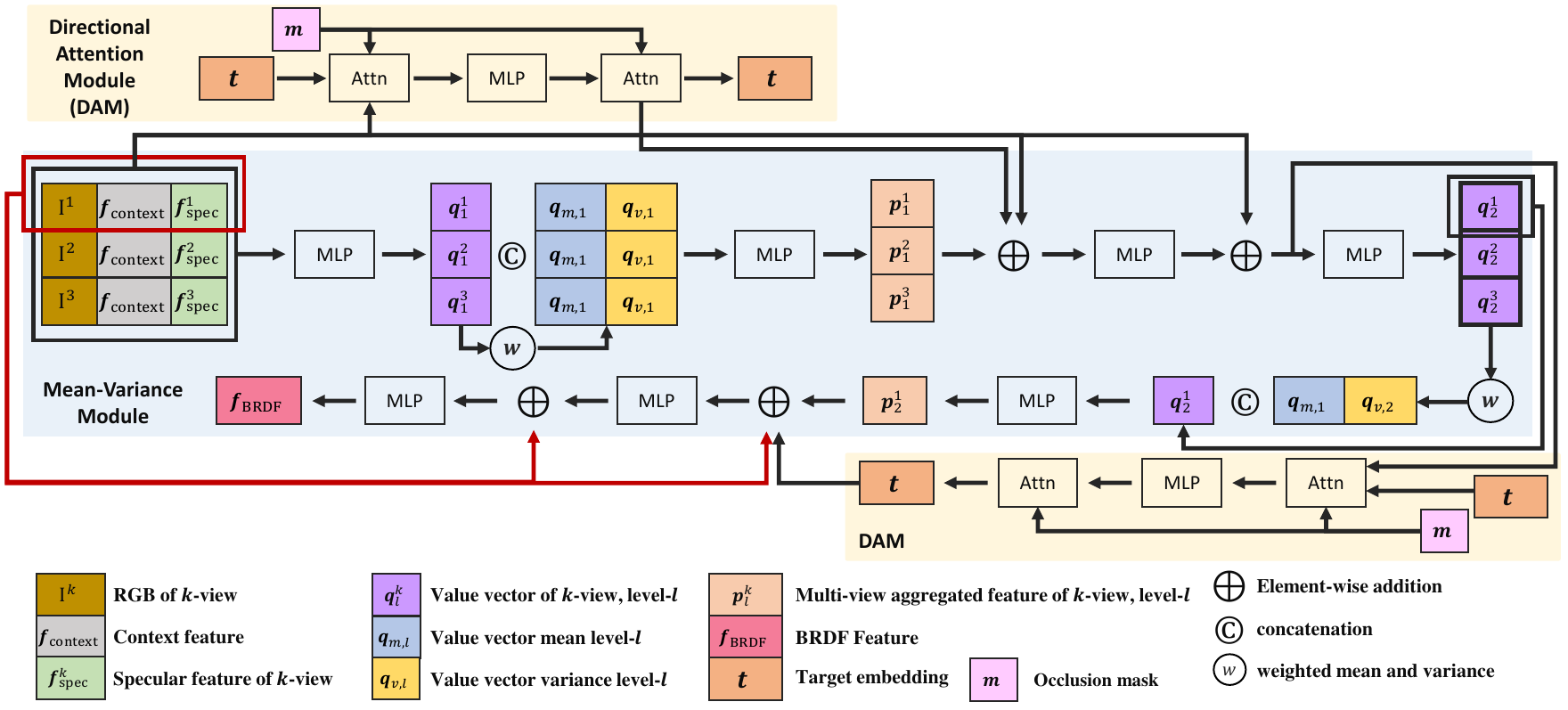}
   \caption{An illustration of MVANet when $K$=3. MVANet creates a value vector by encoding color, context feature, and specular feature, and uses multi-view weights as attention to create multi-view aggregated features. Since our goal is to obtain the BRDF of target-view($1$-view), in level-2, only the value vector of target view is processed. While MAIR used only the Mean-Variance Module, MAIR++ attempted to infer more complex relationships between mulit-views with the DAM.}
   \label{fig:mvanet}
\end{figure*}

\subsection{Stage 1 - Target View Analysis}
Stage 1 consists of three networks: Normal map estimation network (NormalNet), incident direct lighting estimation network (InDLNet), and exitant direct lighting estimation network (ExDLNet). Inspired by recent studies \cite{cis2020, vsg, lighthouse}, MAIR adopts spatially-varying spherical Gaussians (SVSGs)~\cite{cis2020} and volumetric spherical Gaussian (VSG)~\cite{vsg} for the representation of incident lighting and exitant lighting, respectively.

\noindent{\bf Normal map estimation.} 
Unlike single-view-based methods\cite{cis2020, irisformer2022, Li22, vsg} that infer the normal map from the scene context, the normal map ($\mathrm{N}_{t}$) can be directly derived when the depth map $\mathrm{D}_{mvs}$ is available. Thus, NormalNet can have robust performance, especially in real-world environments, where the distribution of image contents and geometry largely differs from that of the training data. In addition, the use of other available information, including the RGB image, the magnitude of the gradient of the depth map ($\nabla{\mathrm{D}_{mvs}} \in \mathbb{R}^{H \times W}$), and the confidence map, can help NormalNet to better handle unreliable depth predictions. Specifically, NormalNet is formulated as follows:
\begin{equation}\label{eqn:eq_normal}
\mathrm{N}_{t} = \text{NormalNet}(\mathrm{I},\mathrm{D}_{mvs}, \nabla\mathrm{D}_{mvs}, \mathrm{C}_{mvs}), \mathrm{N}_t \in \mathbb{R}^{3 \times H \times W}.
\end{equation}
NormalNet has a simple U-Net\cite{unet} structure with six down-up convolution blocks. 

\noindent{\bf Incident direct lighting estimation.} 
Given $\mathrm{I}$, $\mathrm{D}_{mvs}$, $\mathrm{C}_{mvs}$, and $\mathrm{N}_{t}$ obtained using NormalNet, MAIR estimates SVSGs as a lighting representation of incident direct lighting, which is proven to be effective in modeling environment map~\cite{cis2020}. InDLNet is formulated as follows: 

\begin{equation}
\{\boldsymbol{\xi}_{s}\}, \{\lambda_{s} \}, \{\boldsymbol{\eta}_{s} \} = \text{InDLNet}(\mathrm{I}, \mathrm{N}_{t}, \mathrm{D}_{mvs}, \mathrm{C}_{mvs}),
\label{eqn:incident_SGs}
\end{equation}
where $\boldsymbol{\xi}_s \in \mathbb{R}^2 $ is the direction vector outward from the center of the unit sphere, $\lambda_s \in \mathbb{R}$ is sharpness, and $\boldsymbol{\eta}_s \in \mathbb{R}^3$ is the intensity vector. The environment map is then parameterized with $S_D$ SG lobes ${\{\boldsymbol{\xi}_s, \lambda_s, \boldsymbol{\eta}_s\}_{s=1}^{S_D}}$.  For the $s$-th SG, its radiance $\mathcal{G}(\boldsymbol{l})$ in the direction $\boldsymbol{l} \in \mathbb{R}^2$ can be obtained as

\begin{equation}
\mathcal{G}(\boldsymbol{l};\boldsymbol{\eta}_s,\lambda_s,\boldsymbol{\xi}_s)=\boldsymbol{\eta}_{s}e^{\lambda_s(\boldsymbol{l}\cdot \boldsymbol{\xi}_s -1)},
\label{eqn:incident_DL1}
\end{equation}
where $\cdot$ represents the inner product. Using all $S_D$ SG lobes, the incident direct radiance $\mathcal{R}^d_{i}(\boldsymbol{l})$ in the direction $\boldsymbol{l}$  is expressed as
\begin{equation}
\mathcal{R}^d_{i}(\boldsymbol{l}) = \displaystyle\sum_{s=1}^{S_D} \mathcal{G}(\boldsymbol{l};\boldsymbol{\eta}_{s},\lambda_{s},\boldsymbol{\xi}_{s}),
\label{eqn:incident_DL2}
\end{equation}
where $S_D=3$ was found to be sufficient to model much simpler direct lighting. Since the intensity ($\boldsymbol{\eta}_s$) is not related to pixel locations, MAIR uses global intensities ${\boldsymbol{\eta}_s}$ rather than per-pixel intensities. Instead, per-pixel visibility ${\mu_s \in \mathbb{R}}$ was used to account for occlusion. $\mu_s$ is multiplied by ${\boldsymbol{\eta}_s}$ to represent the per-pixel intensities, which is omitted from \cref{eqn:incident_SGs,eqn:incident_DL1,eqn:incident_DL2} for simplicity.

\noindent{\bf Exitant direct lighting estimation.} 
The above environment map alone is insufficient to model lighting in a 3D space. Thus, a voxel-based representation called VSG~\cite{vsg} is adopted to further model exitant direct lighting. ExDLNet estimates the exitant direct lighting volume $\mathrm{V}_\text{DL}$ as
\begin{equation}
\mathrm{V}_\text{DL} = \text{ExDLNet}(\mathrm{I}, \mathrm{N}_{t}, \mathrm{D}_{mvs}, \mathrm{C}_{mvs}), \mathrm{V}_\text{DL} \in \mathbb{R}^{8 \times X \times Y \times Z},
\end{equation}
where $X$, $Y$, and $Z$ are the sizes of the volume. Each voxel in $\mathrm{V}_\text{DL}$ contains opacity $\alpha$ and SG parameters ($\boldsymbol{\eta}, \boldsymbol{\xi}, \lambda$). From VSG, alpha compositing in the direction $\boldsymbol{l}$  allows us to calculate the incident direct radiance $\mathcal{R}^d_{e}(\boldsymbol{l})$ as follows:
\begin{equation}
\mathcal{R}^d_{e}(\boldsymbol{l}) = \displaystyle\sum_{n=1}^{N_R}\prod_{m=1}^{n-1} (1-\alpha_m)\alpha_n \mathcal{G}(-\boldsymbol{l};\boldsymbol{\eta}_{n},\lambda_{n},\boldsymbol{\xi}_{n}),
\label{eqn:exitant_SG}
\end{equation}
where $N_R$ is the number of ray samples, and $\boldsymbol{\eta}_{n}$, $\lambda_{n}$, and $\boldsymbol{\xi}_{n}$ are the SG parameters of the sample. 




\subsection{Stage 2 - Material Estimation}
For BRDF estimation, the specular radiance must be considered. To obtain a specular radiance feature $\boldsymbol{f}_{\text{spec}}$, MAIR uses a SpecNet. The diffuse radiance $\mathrm{I}_d$ and specular radiance $\mathrm{I}_s$ of the microfacet BRDF model~\cite{microfacet} are given as follows:

\begin{equation}
\mathrm{I}_d = \frac{\mathrm{A}}{\pi} \mathrm{S},\quad \mathrm{S} = \int_{\boldsymbol{l}} L(\boldsymbol{l})\mathrm{N}\cdot\boldsymbol{l} \,d\boldsymbol{l},
\label{eqn:diffuse_brdf}
\end{equation}

\begin{equation}
\mathrm{I}_s = \int_{\boldsymbol{l}} L(\boldsymbol{l})\mathcal{B}_s(\boldsymbol{v}, \boldsymbol{l}, \mathrm{N}, \mathrm{R}) \mathrm{N}\cdot\boldsymbol{l} \,d\boldsymbol{l},
\label{eqn:specular_brdf}
\end{equation}
where $\mathrm{A}$ is diffuse albedo, $\mathrm{S}$ is shading, $\mathrm{N}$ is normal, $\boldsymbol{l}$ is lighting direction, $L(\boldsymbol{l})$ is lighting intensity, $\mathcal{B}_s$ is specular BRDF, $\mathrm{R}$ is roughness, and $\boldsymbol{v}$ is viewing direction. Since Eq.~\ref{eqn:specular_brdf} is highly complicated, it is necessary to efficiently encode inputs to make it easier for the network to learn. To this intent, the arguments of $\mathcal{B}_s$ are modified as follows:

\begin{equation}
[\mathcal{F}(\boldsymbol{v},\boldsymbol{h}), (\mathrm{N}\cdot \boldsymbol{h})^2, \mathrm{N}\cdot \boldsymbol{l}, \mathrm{N}\cdot \boldsymbol{v}, \mathrm{R}],
\end{equation}
where $\mathcal{F}$ is the Fresnel equation, and $\boldsymbol{h}$ is the half vector. Then, the lighting of Eq.~\ref{eqn:specular_brdf} is approximated with the SVSGs of Eq.~\ref{eqn:incident_SGs}. 
Since each SG lobe ${\{\boldsymbol{\xi}, \lambda, \boldsymbol{\eta} \}}$ can be thought of as an individual light source, $\boldsymbol{\xi}$, $\boldsymbol{\eta}$, and $\lambda$ can be regarded as $\boldsymbol{l}$, $L(\boldsymbol{l})$, and a parameter to approximate the integral, respectively. Consequently, $\mathrm{I}_s$ can be obtained as follows:

\small
\begin{equation}
\mathrm{I}_s = {\sum\limits_{s=1}^{S_D} g(\mathcal{F}(\boldsymbol{v},\boldsymbol{h}_s), (\mathrm{N}\cdot \boldsymbol{h}_s)^2, \mathrm{N}\cdot \boldsymbol{\xi}_s, \mathrm{N}\cdot \boldsymbol{v}, \boldsymbol{\eta}_s, \lambda_s, \mathrm{R})},
\label{eqn:specular_approx}
\end{equation}
\normalsize
where $g$ is a newly defined function from above reparameterization. Using Eq.~\ref{eqn:specular_approx}, MAIR defines SpecNet as follows:

\begin{multline}\label{eqn:fspec}
\boldsymbol{f}_{\text{spec}}^{k} = \displaystyle\sum_{s=1}^{S_D} m_s\text{SpecNet}(\mathcal{F}(\boldsymbol{v}_k,\boldsymbol{h}_{s,k}), (\mathrm{N}_t\cdot \boldsymbol{h}_{s,k})^2, \\
\mathrm{N}_t \cdot \boldsymbol{\xi}_s, \mathrm{N}_t \cdot \boldsymbol{v}_k, \boldsymbol{\eta}_s, \lambda_s), 
\end{multline}

\begin{equation}\label{eqn:eq_mask}
m_s = \begin{cases}
   1 &\text{if } \lVert \boldsymbol{\eta}_s \rVert_1\mathrm{N}_t \cdot \boldsymbol{\xi}_s > 0,\\
   0 &\text{else, } 
\end{cases}
\end{equation}
where $k$ is $k$-th view. A binary indicator $m_s$ is used to exclude SG lobes from $\mathrm{I}_s$ if the intensity of the light source ($\lVert \boldsymbol{\eta}_s \rVert_1$) is 0 or the dot product of the normal and light axis ($\mathrm{N}_t \cdot \boldsymbol{\xi}_s$) is less than 0. Since SpecNet approximates Eq.~\ref{eqn:specular_brdf} with this physically-motivated encoding, $\boldsymbol{f}_{\text{spec}}$ can include a feature for specular radiance information. In addition to SpecNet, MAIR uses ContextNet to obtain a context feature map: 

\begin{equation}\label{eqn:eq_context}
\boldsymbol{f}_{\text{context}} = \text{ContextNet}(\mathrm{I},\mathrm{D}_{mvs},\mathrm{C}_{mvs},\mathrm{N}_{t}),
\end{equation}
that contains the local context of the scene. All views share $\boldsymbol{f}_{\text{context}}$ of the target view. 


Next, a multi-view aggregation network (MVANet) is used to aggregate $\boldsymbol{f}_\text{spec}$, $\boldsymbol{f}_\text{context}$, and $\mathrm{I}$ across the pixels of all $K$ views, which correspond to the target view pixel considering MVS depths. However, some of these pixel values might have a negative effect if they are from incorrect surfaces due to occlusion. To consider occlusion, MAIR defined the depth projection error in $k$-view as follows.

\begin{equation}
e_k = \mid d_k - z_k \mid, \quad \boldsymbol{e} = (e_1, e_2, ..., e_K) \in \mathbb{R}^K,
\label{eqn:depth_proj_error}
\end{equation}
where $d_k$ is the depth at the pixel position obtained by projecting a point seen from the target view onto $k$-view, and $z_k$ is the distance between the point and the camera center of $k$-view. The depth projection error $\boldsymbol{e}$ is obtained by aggregating $e_k$ from all $K$ views. MAIR uses multi-view weight $\boldsymbol{w}$:

\begin{equation}
\boldsymbol{w} = \frac {\max(-\log(\boldsymbol{e}), 0)} {||\max(-\log(\boldsymbol{e}), 0)||_1}, \boldsymbol{w} \in \mathbb{R}^K, 
\label{eqn:multi_view_weight}
\end{equation}
as weights during the multi-view feature aggregation in MVANet. MVANet first encodes the input for each view to produce a value vector $\boldsymbol{q}$, and produces a mean and variance of $\boldsymbol{q}$ according to $\boldsymbol{w}$\cite{ibrnet}. It is encoded again, producing a multi-view aggregated feature $\boldsymbol{p}$. Since $\boldsymbol{p}$ is created from weighted means and variances, it has multi-view information considering occlusion. This process is repeated once again for the target view. This is the Mean-Variance Module of MVANet in Fig.~\ref{fig:mvanet}. 

Since MVANet exploits only local features, long-range interactions within the image need to be further considered for inverse rendering\cite{irisformer2022}. Thus, MAIR proposes RefineNet for albedo ($\mathrm{A}_m \in \mathbb{R}^{3 \times H \times W}$), roughness($\mathrm{R}_m \in \mathbb{R}^{H \times W}$) estimation using $\boldsymbol{f}_{\text{BRDF}}$ from MVANet.

\begin{equation}\label{eqn:eq_refine}
\mathrm{A}_m, \mathrm{R}_m = \text{RefineNet}(\mathrm{I},\mathrm{D}_{mvs},\mathrm{C}_{mvs},\mathrm{N}_{t}, \boldsymbol{f}_{\text{BRDF}}).
\end{equation}

ContextNet is U-Net with ResNet18\cite{resnet}, SpecNet is MLP with 3 layers, and RefineNet is a shared U-Net encoder and two separate decoders with group normalization.

\subsection{Stage 3 - Lighting Estimation}
In Stage 3, the spatially varying lighting estimation network (SVLNet) infers 3D lighting with direct lighting, geometry, and material. To this end, MAIR creates a visible surface volume ($\mathrm{V}_{\text{vis}} \in \mathbb{R}^{10 \times X \times Y \times Z}$) by reprojecting $\mathrm{I}, \mathrm{N}_{t}, \mathrm{A}_m, \mathrm{R}_m$. For each voxel, let $(u, v)$ and $d$ denote the projected coordinates of the center point and the depth, respectively. Then, the local feature $\boldsymbol{f}_\text{vis} \in \mathbb{R}^{10}$ for each voxel is initialized as follows.

\begin{equation}
\boldsymbol{f}_\text{vis} =\left [\rho \mathrm{I}(u,v), ~\rho \mathrm{N}_{t}(u,v), ~\rho \mathrm{A}_m(u,v), ~\rho \mathrm{R}_m(u,v)\right ], 
\end{equation}
where $\rho = e^{-\mathrm{C}_{mvs}(u,v)(d-\mathrm{D}_{mvs}(u,v))^2}$. Note that the confidence ($\mathrm{C}_{mvs}$) is used to reflect the accuracy of the depth. $\mathrm{V}_{\text{vis}}$ and $\mathrm{V}_\text{DL}$ are fed to SVLNet, producing outputs $\mathrm{V}_\text{SVL}$ representing 3D spatially-varying lighting volume, as follows:

\begin{equation}
\mathrm{V}_\text{SVL} = \text{SVLNet}(\mathrm{V}_\text{DL}, \mathrm{V}_{\text{vis}}), \mathrm{V}_\text{SVL} \in \mathbb{R}^{8 \times X \times Y \times Z}.
\end{equation}

In SVLNet, $\mathrm{V}_{\text{vis}}$ is concatenated with $\mathrm{V}_\text{DL}$ after 2 downsampling and processed with 3D U-Net. 

\begin{figure*}[t!]
  \centering
  \includegraphics[width=\linewidth]{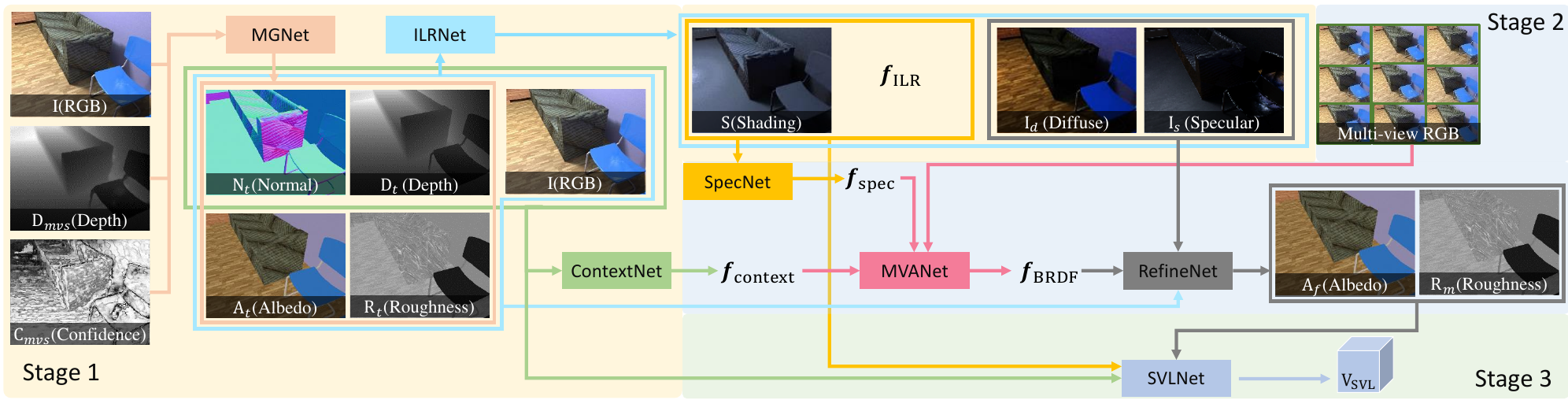}
  \caption{MAIR++'s entire pipeline. Given RGB and MVS depth, MAIR++ performs single-view inverse rendering and infers surface lighting, then further infers it with multi-views. Finally, all information is integrated to calculate the 3D volume lighting.}
   \label{fig:mair++_whole}
\end{figure*}

\section{MAIR++}\label{sec:method2}
This section presents an extended version of MAIR, called MAIR++. The overall pipeline of MAIR++ is shown in Fig.~\ref{fig:mair++_whole}, and its key features are detailed in the following subsections. 

\subsection{Initial Material and Geometry Estimation}
\begin{figure}[ht]
  \centering
  \includegraphics[width=\linewidth]{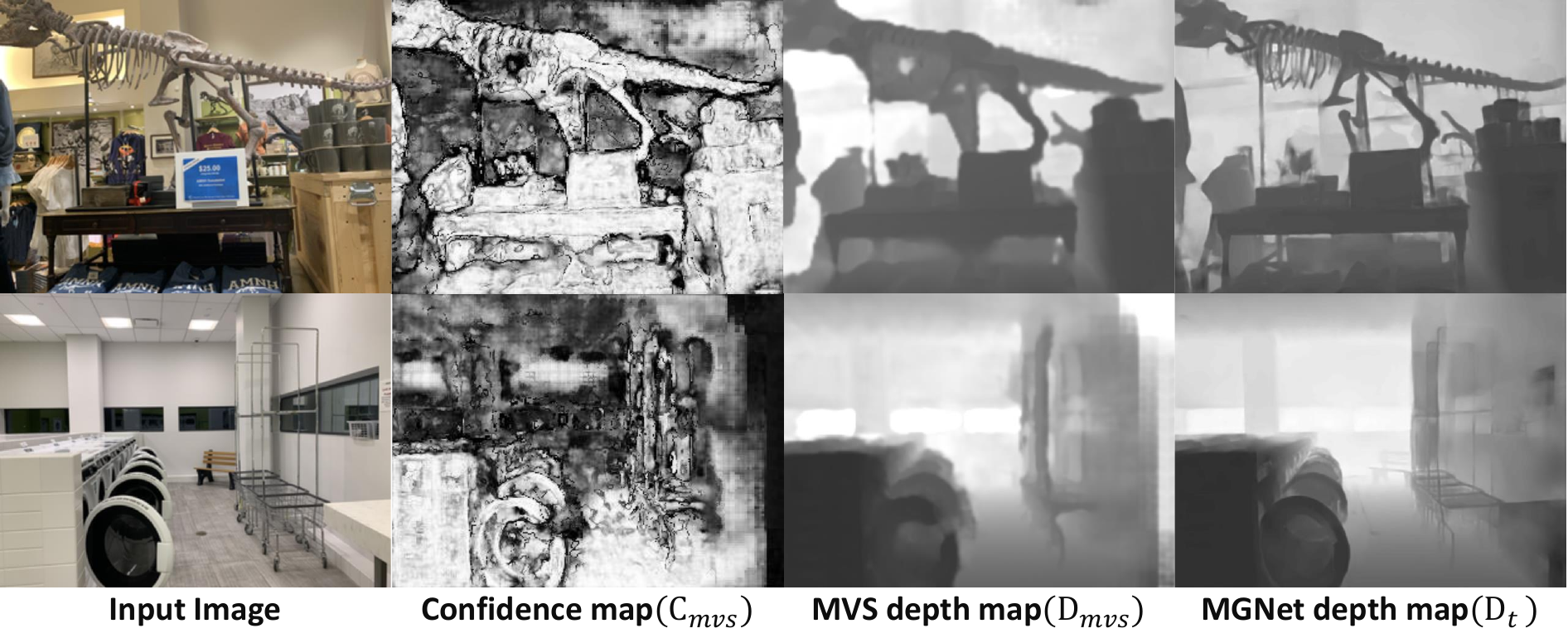}
  \caption{Comparison of depth map quality in unseen real world data.}
  \label{fig:depth_compare}
\end{figure}

One of the limitations of MAIR is that the MVS depth is not accurate enough to be used in complex indoor scenes. To address this limitation, we pay attention to the complementary nature of the two depth prediction methods, the single-view method and the multi-view method~\cite{magnet}. Specifically, in MAIR++, we structure our \textbf{M}aterial \textbf{G}eometry Network (MGNet) as follows:
\begin{equation}\label{eqn:eq_mg}
\mathrm{A}_t, \mathrm{R}_t, \mathrm{N}_t, \mathrm{D}_t = \text{MGNet}(\mathrm{I},\mathrm{D}_{mvs}, \nabla\mathrm{D}_{mvs}, \mathrm{C}_{mvs}).
\end{equation}
Our MGNet can improve depth accuracy by compensating for areas with inaccurate depth predictions using $\mathrm{C}_{mvs}$. Since MGNet uses a normalized depth map as input and output, the depth scale is not determined. Thus, we determine the depth scale from $\mathrm{D}_{mvs}$. Specifically, the depth scale is computed by least squares regression using pixels with a high confidence values ($\mathrm{C}_{mvs} > 0.9$). As shown in Tab.~\ref{tab:depth_acc}, the MGNet depth map shows a higher accuracy than the MVS depth map. Fig.~\ref{fig:depth_compare} shows the results of depth map improvement in an unseen real-world scene. Even though MGNet is trained only on synthetic scenes, it improves $\mathrm{D}_{mvs}$ more accurately on real world scenes. Unlike MAIR, which only predicts the normal map ($\mathrm{N}_t$), we designed MGNet to predict not only the normal map but also the depth, albedo, and roughness maps, denoted as $\mathrm{D}_t$, $\mathrm{A}_t$, and $\mathrm{R}_t$ respectively. These additional maps provide valuable initial values for the later stages of inverse rendering.

\noindent{\bf CNN architecture.} Our MGNet can be implemented as single-view-based methods~\cite{cis2020, zhu2022montecarlo} while taking the MVS depth as an additional input. Through experiments, we found that a DenseNet-based architecture~\cite{zhu2022montecarlo} is more effective in inferring material and geometry than a simple U-Net~\cite{cis2020}, as shown in Tab.~\ref{tab:compare}. Thus, we design our MGNet using DenseNet121 with a shared encoder and four separate decoders. 

\begin{table}[ht]
\footnotesize
\centering
\resizebox{\linewidth}{!}{
\begin{tabular}{|l|c|c|c|}
\hline
Depth & \multicolumn{1}{|c|}{$\mathrm{D}_{mvs}$} & \multicolumn{1}{|c|}{$\mathrm{D}_t$(GT scale)} & \multicolumn{1}{|c|}{$\mathrm{D}_t$($\mathrm{D}_{mvs}$ scale)}\\
\hline
MAE($\times 10^{-2}$) & 2.3135 & 0.6339 &0.6788\\
\hline
\end{tabular}
}
\caption{MAE(Mean Absolute Error) for depth maps of OpenRooms FF test dataset.}
\label{tab:depth_acc}
\end{table}

\begin{figure}[ht]
  \centering
  \includegraphics[width=\linewidth]{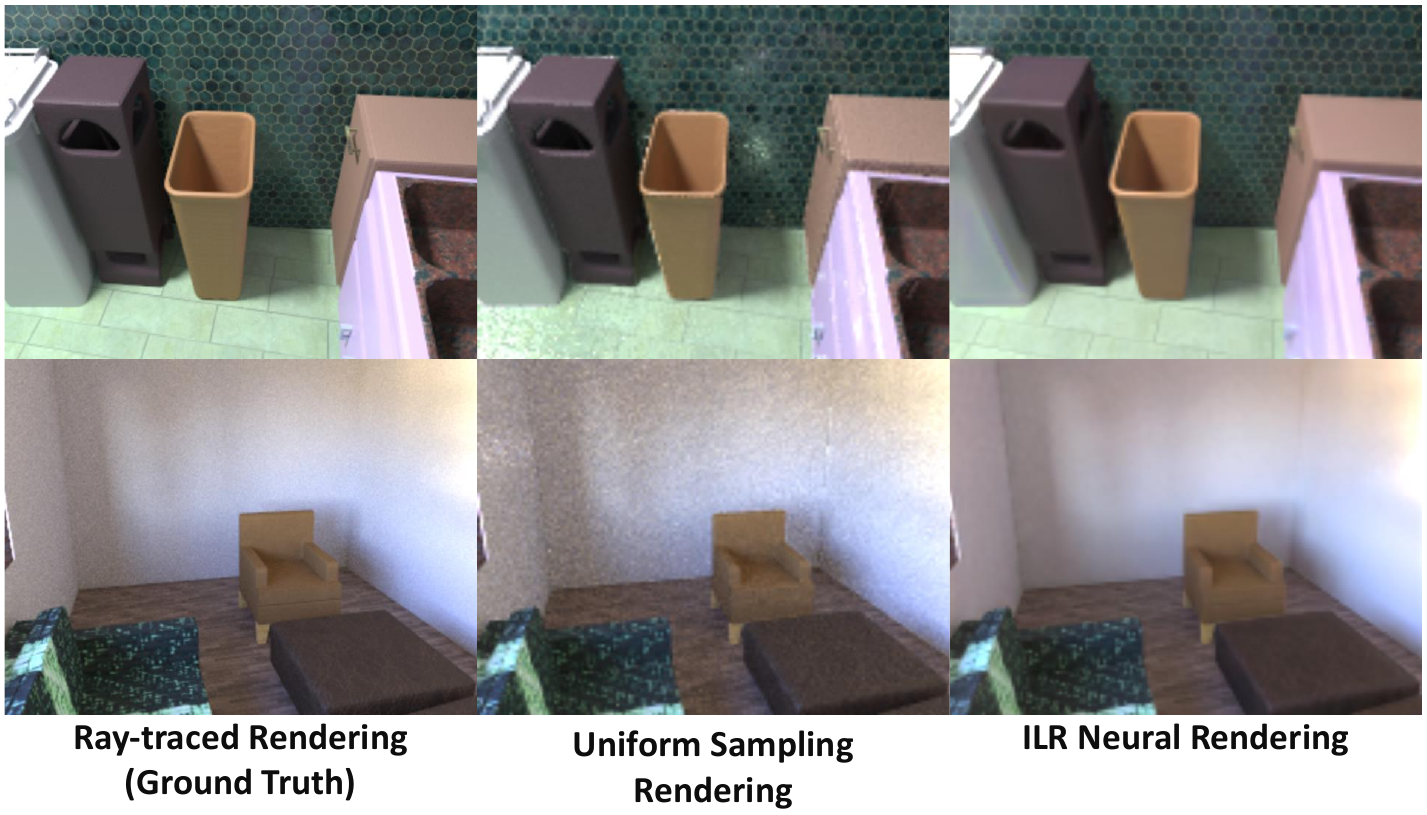}
  \caption{Comparison of rendering methods. Even though our ILR uses predictions, it provides more realistic rendering than uniform sampling rendering using ground truth.}
  \label{fig:rendering_compare}
\end{figure}

\subsection{Implicit Lighting Representation}
\begin{figure}[ht]
  \centering
  \includegraphics[width=\linewidth]{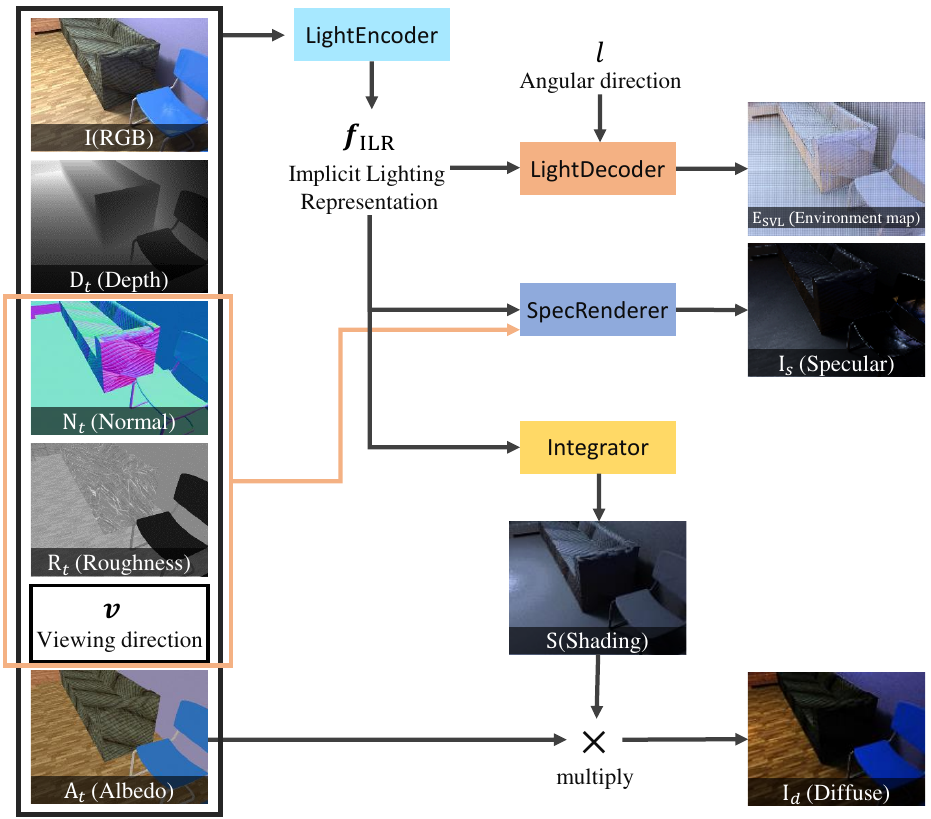}
   \caption{Overview of ILRNet. LightEncoder infers surface lighting, $\boldsymbol{f}_\text{ILR}$, from the initial inverse rendering results. This can be converted into an environment map with LightDecoder, and the image can be rendered realistically through Integrator and SpecRenderer.}
   \label{fig:ilr}
\end{figure}

In inferring BRDF, MAIR focused on direct lighting, which directly affects specular radiance, allowing it to estimate materials with few parameters. However, direct lighting cannot accurately estimate the material due to its limited representation ability. A straightforward extension is to replace MAIR's direct lighting with spatially-varying lighting. This works to some extent; however, we found that SVSGs do not sufficiently capture the rich information of light. More importantly, SVSG-based rendering samples incident lighting uniformly in the $\theta$-$\phi$ space, making realistic rendering difficult. This challenge persists even for ground truth lighting. Because OpenRooms' per-pixel environment map is uniformly sampled, it cannot reproduce images realistically, as shown in Fig.~\ref{fig:rendering_compare}. Uniform sampling rendering exhibits significant noise and artifacts even when using ground truth material, geometry, and lighting. To overcome these limitations, we propose a new lighting representation called ILR that encodes lighting as a neural feature vector. Specifically, ILR represents the incident lighting of each pixel as a feature vector. This can be converted to an environment map through a LightDecoder, to shading through an Integrator, and to rendering specular radiance through a SpecRenderer. The LightEncoder is formulated as follows:
\begin{equation}\label{eqn:eq_lightencoder}
\boldsymbol{f}_\text{ILR} = \text{LightEncoder}(\mathrm{I}, \mathrm{D}_{t}, \mathrm{N}_{t}, \mathrm{A}_{t}, \mathrm{R}_{t}, \mathrm{N}_{t} \cdot \boldsymbol{v}),
\end{equation}
where $\boldsymbol{v}$ is the viewing direction of the target image. $\mathrm{N}_{t} \cdot \boldsymbol{v}$ is used as input to recognize specular radiance in the input image. LightEncoder is implemented as a DenseNet121-based encoder-decoder. The other networks can be expressed as:
\begin{equation}\label{eqn:eq_lightdecoder}
\mathcal{R}_{i}(\boldsymbol{l}) = \text{LightDecoder}(\boldsymbol{f}_\text{ILR}, \gamma(\boldsymbol{l})),
\end{equation}
\begin{equation}\label{eqn:eq_integrator}
\mathrm{S} = \text{Integrator}(\boldsymbol{f}_\text{ILR}),
\end{equation}
\begin{equation}\label{eqn:eq_specrenderer}
\mathrm{I}_s = \text{SpecRenderer}(\boldsymbol{f}_\text{ILR}, \mathrm{R}, \gamma(\boldsymbol{r}), \mathrm{N}_{t} \cdot \boldsymbol{v}),
\end{equation}
where $\boldsymbol{l}$ is the angular direction of the environment map, $\gamma(\cdot)$ is the frequency positional encoding~\cite{nerf} that helps MLP represent high frequency information, $\mathcal{R}_{i}$ is the incident lighting, and $\boldsymbol{r}$ is the reflection direction. Unlike LightEncoder, these three networks are all implemented as MLPs. (That is, they infer at the pixel-level.) Given $\boldsymbol{f}_\text{ILR}$ and $\boldsymbol{l}$, LightDecoder decodes it into an environment map. Integrator calculates $\mathrm{S}$ (shading) by integrating $\boldsymbol{f}_\text{ILR}$. SpecRenderer takes as input $\boldsymbol{f}_\text{ILR}$, $\mathrm{R}$, $\boldsymbol{r}$, and $\mathrm{N}_{t} \cdot \boldsymbol{v}$ needed for specular radiance ($\mathrm{I}_s$). Integrator and SpecRenderer can render images faster and more realistically than rendering by directly decoding $\boldsymbol{f}_\text{ILR}$ into an environment map. We name these four networks together ILRNet, and they are trained simultaneously. The overview of ILRNet is shown in Fig.~\ref{fig:ilr}. 

\begin{figure}[ht]
  \centering
  \includegraphics[width=\linewidth]{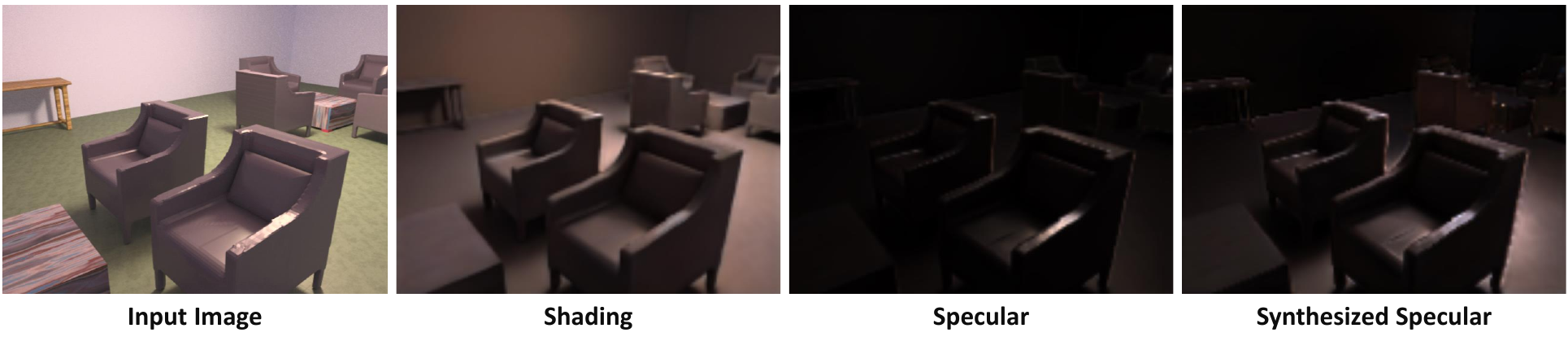}
   \caption{ILRNet rendering example. Synthetized specular is an image where viewing direction and roughness have been changed.}
   \label{fig:ilr_example}
\end{figure}

\noindent{\bf Training.} Given that ILRNet's renderer (Integrator and SpecRenderer) is also expected to edit materials, it is crucial to prevent the renderer from converging to a local minimum, such as simply reproducing the input RGB values. To address this issue, ILRNet's renderer is divided into diffuse and specular components. Additionally, during training, since MGNet's $\mathrm{A}_t$, $\mathrm{R}_t$, and $\mathrm{N}_t$ may be inaccurate, we render using their ground truth, $\widetilde{\mathrm{A}}$, $\widetilde{\mathrm{R}}$, and $\widetilde{\mathrm{N}}$. In other words, the renderer is trained to focus primarily on lighting regardless of material. This helps prevent the renderer from outputting RGB as is, and also helps with material estimation later. We also experimented with using random albedo and roughness, but using only the ground truth was found to be sufficient. Fig.~\ref{fig:ilr_example} shows an example of ILRNet rendering. ILRNet not only separates the shading and specular components of images well, but can also render arbitrary specular images.

After being trained, ILRNet is used in Stage 2 and Stage 3 of MAIR++. First, in Stage 2, $\boldsymbol{f}_\text{ILR}$ and $\mathrm{S}$ replace MAIR's incident direct lighting parameters ($\{\boldsymbol{\xi}, \lambda, \boldsymbol{\eta}\}$). Since each of $\{\boldsymbol{\xi}, \lambda$, and $\boldsymbol{\eta}\}$ in MAIR has a definite physical meaning, physically motivated encodings (\ref{eqn:fspec}) are used in SpecNet to deal with them. On the other hand, in MAIR++, incident lighting is used as follows.

\begin{equation}\label{eqn:eq_specnet_mair++}
\boldsymbol{f}_{\text{spec}}^{k} = \text{SpecNet}(\boldsymbol{f}_\text{ILR}, \gamma(\boldsymbol{r_k}), \mathrm{N}_{t} \cdot \boldsymbol{v}_k, \mathrm{S}).
\end{equation}

SpecNet, like MAIR, is a three-layer MLP. Additionally, $\mathrm{I}_d$ and $\mathrm{I}_s$ rendered by ILRNet can help RefineNet predict albedo through the AFM.

\subsection{Directional Attention Module}
\begin{figure}[ht]
  \centering
  \includegraphics[width=\linewidth]{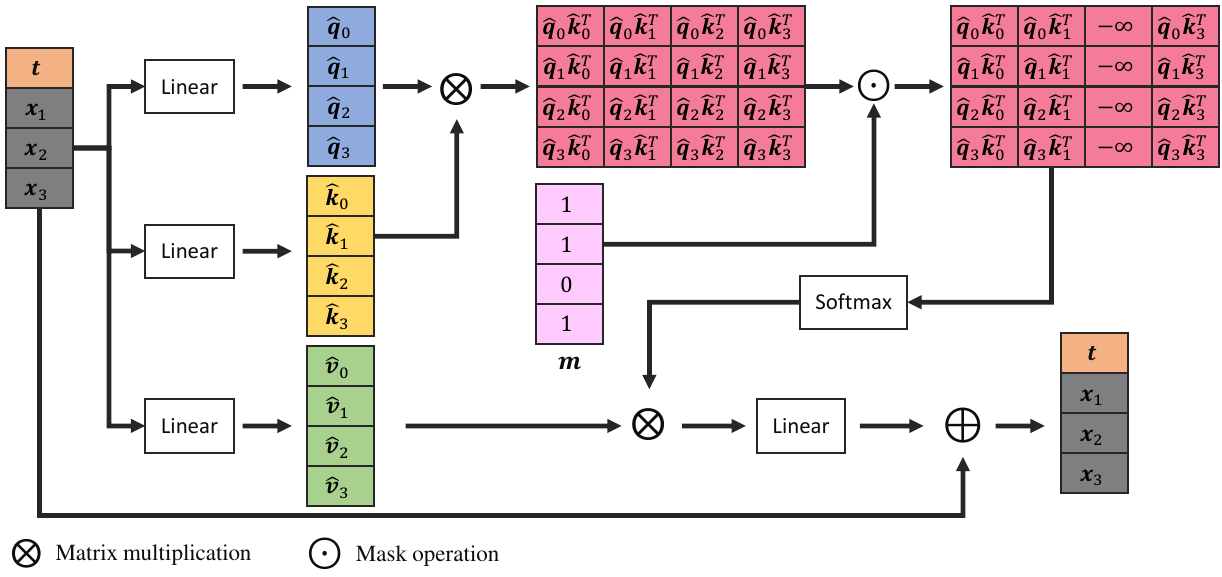}
   \caption{Example of attention layer when m = [1,1,0,1]. $\hat{q}$, $\hat{k}$, and $\hat{v}$ mean query, key, and value, respectively, and $\boldsymbol{x}_k$ is the input token for $k$-view.}
   \label{fig:attn}
\end{figure}

MVANet based on the mean-variance module showed that it could infer the material by focusing on the amount of change in RGB according to the viewing direction. However, mean and variance have limitations in inferring the complex correlation between RGB and BRDF, which prevents more accurate material prediction. To infer more complex relationships of RGB variation between multi-views, we propose the Directional Attention Module (DAM). The DAM treats the features of each view as tokens and infers their directional relationships. To consider further occlusion in multi-views, we design two attention modules. One is the masked attention and the other is the weighted attention. Based on the depth projection error ($\boldsymbol{e}$ of Eq.~\ref{eqn:depth_proj_error}), the mask ($\boldsymbol{m}$) of the masked attention is defined as follows. 

\begin{equation}
m_k = \begin{cases}
   1 &\text{if } e_k < c_{th}\ \text{for} \ k=1, ..., K \\
   0 &\text{else, }              
\end{cases},
\end{equation}

\begin{equation}
\boldsymbol{m} = (1, m_1, m_2, ..., m_K) \in \mathbb{R}^{K+1},    
\label{eqn:multi_view_mask}
\end{equation}
where $c_{th}$ is the threshold for the mask and in our experiments we used 0.05 m. Similar to the mask of classical self attention~\cite{transformer}, we make the attention score of the part where $m_k$ is $0$ to $-\infty$ and force the value vector for it to be ignored. However, unlike the previous look-ahead mask that rejects the use of future data, we use a mask to remove specific data. Specifically, the first element of $\boldsymbol{m}$ is the mask for the target embedding $\boldsymbol{t}$ and is always set as 1. $\boldsymbol{t}$ has a similar role to ViT~\cite{vit}'s class token and is used to collect multi-view information that is inferred through the attention layer. Fig.~\ref{fig:attn} illustrates the attention layer when $K=3$ and $\boldsymbol{m}=[1,1,0,1]$. 

In weighted attention, we multiply the attention score by multi-view weight ($\boldsymbol{w}$ of Eq.~\ref{eqn:multi_view_weight}) and then applied L1 normalization. Both of these attention methods allow the directional attention module to properly handle occlusion (see Tab.~\ref{tab:abl_BRDF_MAIR++}), and unless otherwise noted, we use masked attention in this paper.

One of the main reasons for the success of classical self-attention~\cite{transformer} is the use of positional embedding, which refers to the location of each word. Although the order of each view is not important in our multi-view setup, we attempt to provide information of each viewing direction to the input token ($\boldsymbol{x}_k$) in the same way as positional embedding. Therefore, we construct $\boldsymbol{x}_k$ as follows:

\begin{equation}\label{eqn:da_pe}
\boldsymbol{x}_{k}= \boldsymbol{f}_{\text{spec}}^{k} + concat(\mathrm{I}^k, \boldsymbol{f}_{\text{context}}).
\end{equation}
This construction is more effective than concatenating all $\boldsymbol{f}_{\text{spec}}^{k}$, $\mathrm{I}^k$, and $\boldsymbol{f}_{\text{context}}$ in training the DAM. Although the DAM can be used alone, we found that the mean-variance module and the DAM are complementary (see Tab.~\ref{tab:abl_BRDF_MAIR++}), and using them together lead to the highest accuracy. Thus, we place these two modules in parallel to allow MVANet to understand the basic relationship between multi-views through the mean-variance module and to understand more complex relationships through the DAM. See Fig.~\ref{fig:mvanet}. 

\subsection{Albedo Fusion Module}
\begin{figure}[ht]
  \centering
  \includegraphics[width=\linewidth]{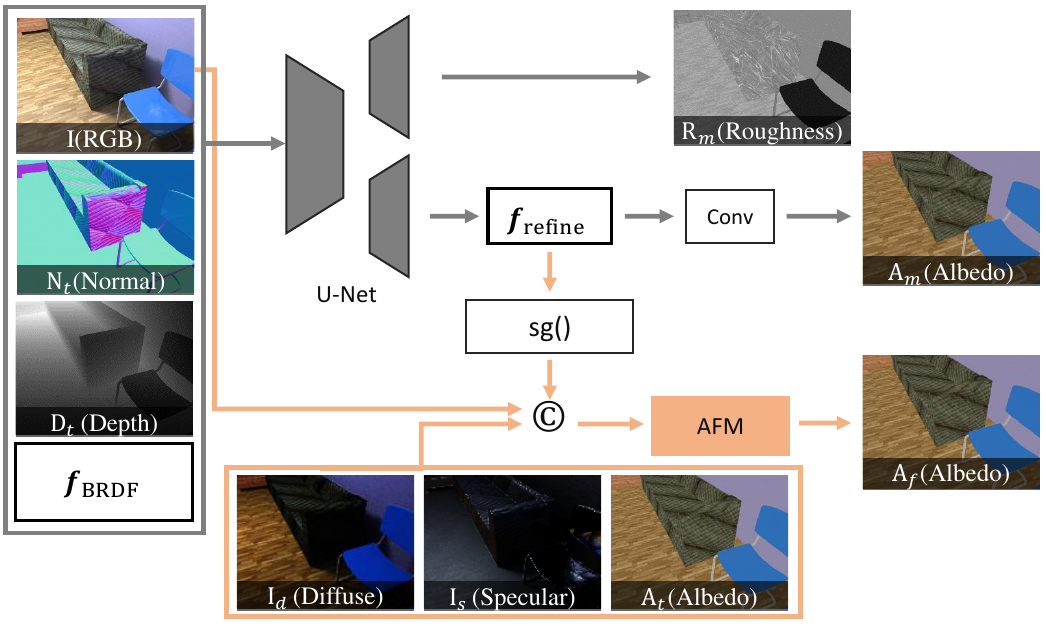}
   \caption{An illustration of RefineNet and AFM. The sg operator is needed to prevent RefineNet from falling into a local minimum. After Stage 2, $\mathrm{A}_f$ is used, but the same loss is also applied to $\mathrm{A}_m$ for training.}
   \label{fig:refine}
\end{figure}

In general, when predicting materials in Stage 2, it may be helpful to use the diffuse albedo map ($\mathrm{A}_t$) and roughness map ($\mathrm{R}_t$) predicted by MGNet. The naive solution is to feed $\mathrm{A}_t$ and $\mathrm{R}_t$ together to RefineNet. However, in this case, RefineNet can fall into a local minimum, and ContextNet and MVANet cannot be trained at all. We speculate that this is because RefineNet relied heavily on $\mathrm{A}_t$ and $\mathrm{R}_t$. Therefore, we add the Albedo Fusion Module (AFM) to the last layer of the albedo decoder of RefineNet. The AFM is trained to appropriately take multi-view albedo ($\mathrm{A}_m$) and single-view albedo ($\mathrm{A}_t$) based on single-view rendering results $\mathrm{I}_d$ and $\mathrm{I}_s$. The inputs and outputs of the AFM are as follows:
\begin{equation}
\mathrm{A}_f = \text{AFM}(\mathrm{I}, \mathrm{A}_t, \text{sg}(\boldsymbol{f}_\text{refine}), \mathrm{I}_d, \mathrm{I}_s),
\end{equation}
where $\boldsymbol{f}_\text{refine}$ is the feature map of the last layer and sg($\cdot$) is stop gradient operator. It is important to apply the stop gradient operator to $\boldsymbol{f}_\text{refine}$ to prevent MVANet from falling into the local minimum. Because $\mathrm{I}_d$ and $\mathrm{I}_s$ are rendered from $\mathrm{A}_t$ and $\mathrm{R}_t$, the AFM can appropriately select $\mathrm{A}_t$ and $\boldsymbol{f}_\text{refine}$ by comparing $\mathrm{I}$, $\mathrm{I}_d$, and $\mathrm{I}_s$. Last, the albedo map $\mathrm{A}_f$ is produced by the AFM, but for MVANet training, we train $\mathrm{A}_m$ with the same loss. Also, because $\mathrm{R}_t$ is significantly inferior to $\mathrm{R}_m$, applying the AFM to the roughness map do not have much effect (See Tab.~\ref{tab:abl_fusion}). See Fig.~\ref{fig:refine} for detailed explanation. The AFM consists of two small convolutional layers.

\begin{table}[htb]
\centering
\begin{tabular}{|c|c|c|c|}
\hline
(si)-MSE ($\times 10^{-2}$) & \begin{tabular}[c]{@{}c@{}}Target view\\ ($\mathrm{A}_t, \mathrm{R}_t$)\end{tabular} & \begin{tabular}[c]{@{}c@{}}Multi-view\\ ($\mathrm{A}_m, \mathrm{R}_m$)\end{tabular} & \multicolumn{1}{c|}{\begin{tabular}[c]{@{}c@{}}Fusion\\ ($\mathrm{A}_f, \mathrm{R}_f$)\end{tabular}} \\ \hline
Albedo($\mathrm{A}\downarrow$)        & 0.5105  & 0.507 & \textbf{0.478}  \\ \hline
Roughness($\mathrm{R}\downarrow$)     & 5.527   & \textbf{2.497} & 5.525  \\ \hline
\end{tabular}
\caption{RefineNet’s AFM experiment results. For the roughness map, the AFM had no effect.}
\label{tab:abl_fusion}
\end{table}

\subsection{3D Lighting Estimation} In MAIR, we trained a relatively simple Exitant direct lighting volume ($\mathrm{V}_\text{DL}$) in advance to help train SVLNet. If we follow this procedure in MAIR++, we need to train $\mathrm{V}_\text{DL}$ expressed as $\boldsymbol{f}_\text{ILR}$. However, this training requires large resources. Instead, we found that simply including $\boldsymbol{f}_\text{ILR}$, $\mathrm{S}$, and $\mathrm{I}_s$ in the visible surface volume ($\mathrm{V}_{\text{vis}}$) can replace $\mathrm{V}_\text{DL}$. In MAIR++, the local feature $\boldsymbol{f}_\text{vis}$ of $\mathrm{V}_{\text{vis}}$ is as follows: 

\begin{multline}
\boldsymbol{f}_\text{vis} = [\rho \mathrm{I}(u,v), ~\rho \mathrm{N}_{t}(u,v), ~\rho \mathrm{A}_f(u,v), ~\rho \mathrm{R}_m(u,v), \\
~\rho \mathrm{I}_s(u,v), ~\rho \mathrm{S}(u,v), ~\rho \boldsymbol{f}_\text{ILR}(u,v) ],     
\end{multline}
where $\rho = e^{\frac{-(d-\mathrm{D}_t(u,v))^2}{2\sigma_\text{d}^2}}$ and $\sigma_\text{d}$ is a hyper-parameter representing uncertainty, and in the experiment we used $0.15$. Note that unlike MAIR, $\mathrm{C}_{mvs}$ is no longer used. In MAIR, SVLNet filled $\mathrm{V}_\text{DL}$ with lighting outside the field of view. In MAIR++, zero-volume $\mathrm{V}_\text{0}$ replaces this. 

\begin{equation}
\mathrm{V}_\text{SVL} = \text{SVLNet}(\mathrm{V}_\text{0}, \mathrm{V}_{\text{vis}}), \mathrm{V}_\text{SVL} \in \mathbb{R}^{8 \times X \times Y \times Z}.
\end{equation}

However, for memory efficiency, we down-sampled $\mathrm{V}_{\text{vis}}$ once and then concatenate $\mathrm{V}_\text{0}$ with $\mathrm{V}_{\text{vis}}$. Additionally, we found that compositing the alpha before calculating the spherical Gaussians radiance is more efficient in terms of memory usage, computation speed, and overall performance, as shown in Tab.~\ref{tab:sg_order}. Specifically, (\ref{eqn:exitant_SG}) is transformed as follows:

\begin{equation}
\mathcal{R}^d_{e}(\boldsymbol{l}) = \mathcal{G}\left(-\boldsymbol{l}; \sum_{n=1}^{N_R} \omega_n \boldsymbol{\eta}_n, \sum_{n=1}^{N_R} \omega_n \lambda_n, \sum_{n=1}^{N_R} \omega_n \boldsymbol{\xi}_n \right)
\label{eqn:sg_after}
\end{equation}

\noindent where
\begin{equation}
\omega_n = \prod_{m=1}^{n-1} (1-\alpha_m)\alpha_n
\end{equation}

\begin{table}[ht]
\centering
\begin{tabular}{|l|c|c|c|}
\hline
Metric &{SG before(Eq.~\ref{eqn:exitant_SG})} & {SG after(Eq.~\ref{eqn:sg_after})}\\
\hline
E($\times 10^{-2}$) & 11.73 & 11.61 \\ \hline
Seconds / iteration & 1.56  & 1.45  \\ \hline
GPU Memory(GB)      & 24    & 20    \\ \hline
\end{tabular}
\caption{Experimental results different in operation order.} 
\label{tab:sg_order}
\end{table}

\section{Implementation Details}

\begin{figure*}[ht]
  \centering
  \includegraphics[width=\linewidth]{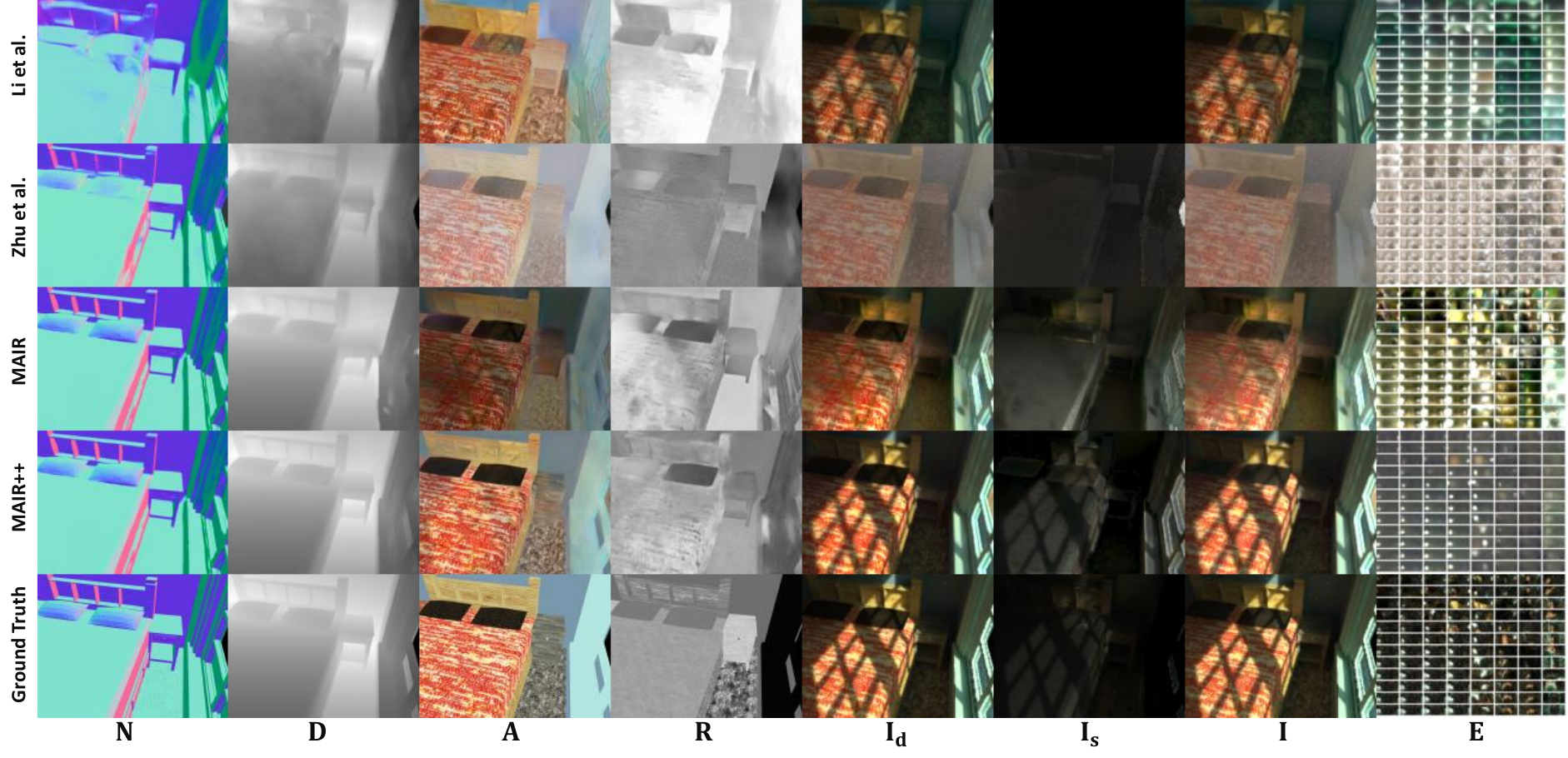}
   \caption{Inverse rendering results for OpenRooms FF test data. MAIR and MAIR++ predicted the geometry of the bed and pillow more accurately than single-view methods, and MAIR++ was the only one to successfully remove the strong shadows of the bed.} 
   \label{fig:ir_syn}
\end{figure*}

\begin{figure*}[ht]
  \centering
  \includegraphics[width=\linewidth]{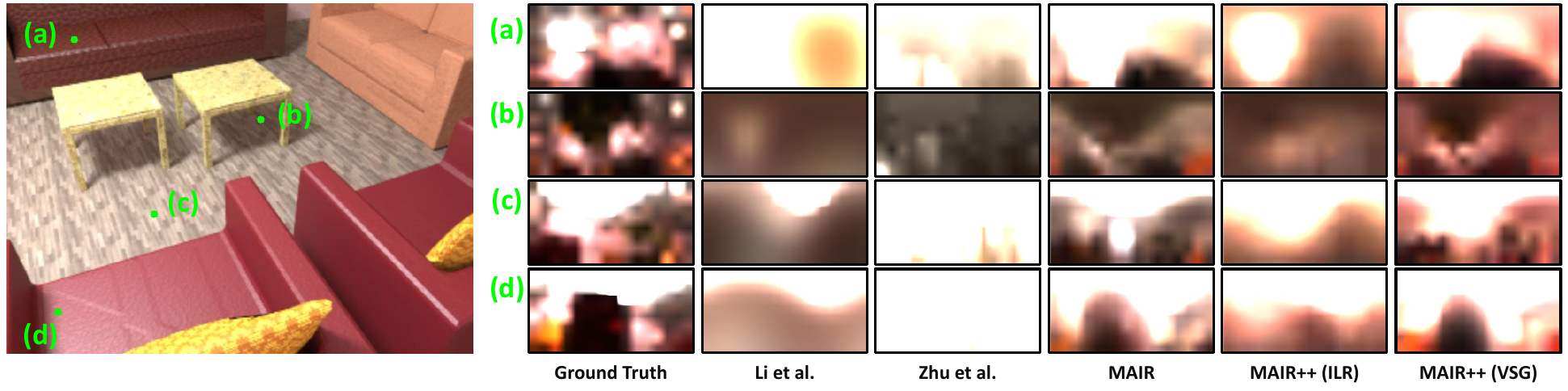}
   \caption{Qualitative analysis of per-pixel environment map in OpenRooms test scene.} 
   \label{fig:eval_light}
\end{figure*}

\subsection{Training strategy and losses} We used nine images ($K$=9) in our experiments and used the nonlinear transformation~\cite{cis2020} for the prediction of HDR lighting. We trained each stage separately, since the ground truth required for each stage is available in OpenRooms FF. We trained lighting with only 2D per-pixel ground truth lighting, but our VSG can represent lighting well in any 3D space (see Fig.~\ref{fig:oi_all}.). Our experiments were carried out with 8 NVIDIA RTX A5000 (24GB). In training, we use Adam optimizer and the binary mask image $(\mathrm{M}_o, \mathrm{M}_l)$. $\mathrm{M}_o \in \mathbb{R}^{H \times W}$ is mask on pixels of valid materials, and $\mathrm{M}_l \in \mathbb{R}^{H \times W}$ is mask on pixels of valid materials and area lighting. The binary mask image is included in the OpenRooms FF and is used only for training. First, we define masked L1 angular error function ($g_1$), masked MSE function ($g_2$), masked scale invariant MSE function ($g_3$), masked $\log$ space MSE function ($g_4$), masked scale invariant $\log$ space MSE function ($g_5$), and regularization function ($g_6$) as follows.

\small
\begin{align}
g_1(A, B, M) = ||(\cos^{-1}(A \odot B)) \otimes M||_1, \\
g_2(A, B, M) = ||{(A - B) \otimes M}||_2^2, \\
g_3(A, B, M) = ||{(A - \tau B) \otimes M}||_2^2, \\
g_4(A, B, M) = ||(\text{log}(A+1) - \text{log}(B+1)) \otimes M) ||_2^2, \\
g_5(A, B, M) = ||(\text{log}(A+1) - \text{log}(\tau B+1)) \otimes M) ||_2^2, \\
g_6(A) = -A\log(A),
\end{align}
\normalsize
where $\odot$ is element-wise dot product, $\otimes$ is element-wise multiplication, and $\tau$ is the scale obtained by least square regression between A and B. $\widetilde{\cdot}$ indicates the ground truth of the element and $\beta$ is the weight of loss. 

In stage 1, the loss function of MGNet is as follows:
\small
\begin{multline} 
\mathcal{L}_{\text{MG}} = \beta_1 g_1(\mathrm{N}_t, \widetilde{\mathrm{N}}, \mathrm{M}_l) 
+ \beta_2 g_2(\mathrm{N}_t, \widetilde{\mathrm{N}}, \mathrm{M}_l)
+ \beta_3 g_5(\mathrm{D}_t, \widetilde{\mathrm{D}}, \mathrm{M}_l) \\ 
+ \beta_4 g_3(\mathrm{A}_t, \widetilde{\mathrm{A}}, \mathrm{M}_o) 
+ \beta_5 g_2(\mathrm{R}_t, \widetilde{\mathrm{R}}, \mathrm{M}_o)
\end{multline}
\normalsize



The loss of ILRNet is as follows:

\small
\begin{align}\label{eqn:eq_lossILR}
\mathcal{L}_{\text{ILR}} = & (1-\alpha) \beta_1 g_4(\mathrm{E}, \widetilde{\mathrm{E}}, \mathrm{M}_o) + 
\alpha \beta_1 g_5(\mathrm{E}, \widetilde{\mathrm{E}}, \mathrm{M}_o) \nonumber \\
& + \beta_2 g_6(\dot{\mathrm{E}}) + 
\beta_3 g_5(\mathrm{I}_d, \widetilde{\mathrm{I}}_d, \mathrm{M}_o) \nonumber + \beta_4 g_5(\mathrm{I}_s, \widetilde{\mathrm{I}}_s, \mathrm{M}_o), 
\end{align}
\normalsize

where $\alpha$ is 0 at 0 training step and linearly increases to 1 at 10000 step, $\mathrm{E}$ is the per-pixel spatially-varying environment map, and $\dot{\mathrm{E}}$ is the output value of the network before non-linear HDR transformation. When ILRNet was trained with scale-invariant loss at half-precision floating point (16-bits), the network was overly dependent on the ground truth scale and the output value converged to 0. To prevent this, we used $\alpha$ to train the lighting scale for the first 10,000 steps, and also added a $\dot{\mathrm{E}}$ term to prevent the network's output from converging to 0.

In stage2, the loss function is as follows. 

\small
\begin{multline}
\mathcal{L}_{\text{BRDF}} = 
\beta_1 g_3(\mathrm{A}_m, \widetilde{\mathrm{A}}, \mathrm{M}_o) + 
\beta_1 g_3(\mathrm{A}_f, \widetilde{\mathrm{A}}, \mathrm{M}_o) + \\
\beta_2 g_2(\mathrm{R}_m, \widetilde{\mathrm{R}}, \mathrm{M}_o).
\end{multline}
\normalsize

In stage3, the loss function is as follows. 
\small
\begin{equation}
\mathcal{L}_{\text{SVL}} = \beta_1 g_5(\mathrm{E}, \widetilde{\mathrm{E}}, \mathrm{M}_o) + \beta_2 g_6(\alpha).
\end{equation} 
\normalsize
In SVLNet, the resolution of the $\mathrm{V}_{\text{vis}}$ and $\mathrm{V}_\text{SVL}$ is $128^3$. SVLNet uses instance normalization(IN). SVLNet needs a lot of memory when training, so we sampled only 4000 environment maps out of the total per-pixel environment maps.

\begin{figure*}[ht]
  \centering
  \includegraphics[width=\linewidth]{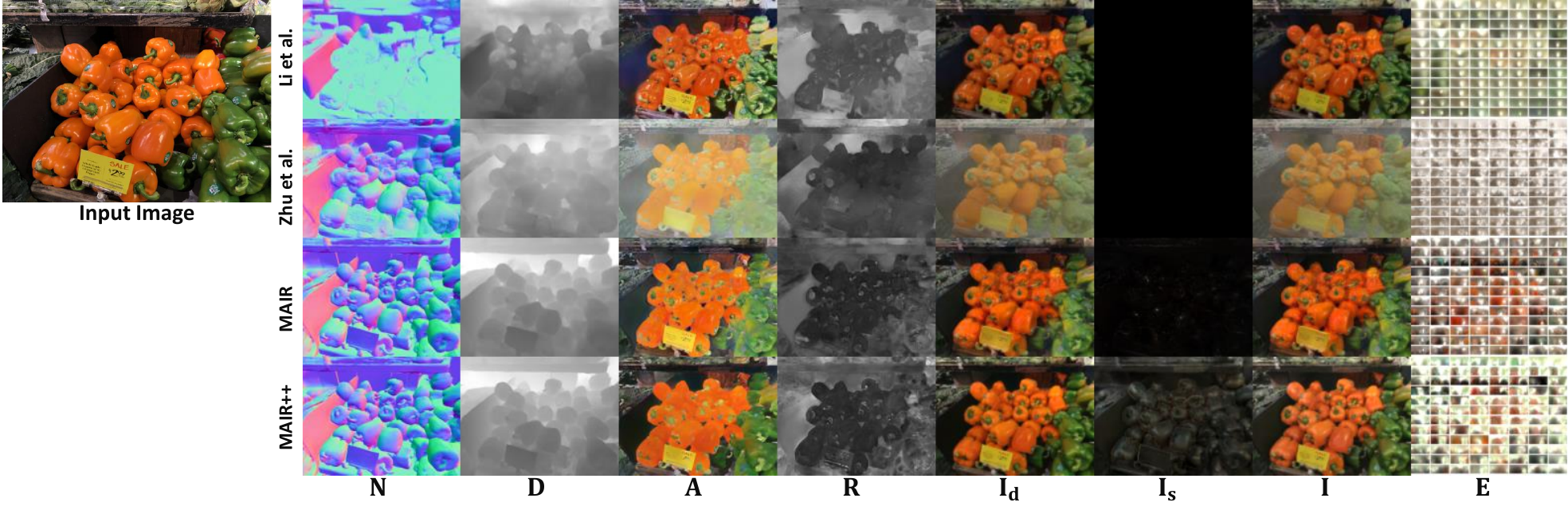}
   \caption{Inverse rendering results for unseen real-world data. Single view methods had difficulty inferring geometry from unseen real world data, where it was difficult to infer context information. On the other hand, MAIR and MAIR++ were able to obtain plausible geometry, and MAIR++ removed specularity from albedo more completely than MAIR.} 
   \label{fig:ir_real}
\end{figure*}

\section{Experiments}
\label{sec:experiments}
We compare our method qualitatively and quantitatively with single-view-based methods~\cite{cis2020, zhu2022montecarlo} and our previous work MAIR. We also performed a qualitative performance evaluation on the real-world dataset~\cite{ibrnet}. Furthermore, we demonstrate the effectiveness of ILR by showing that materials can be realistically edited with ILR. Finally, we demonstrate our realistic 3D spatially-varying lighting through virtual object insertion. 

\subsection{Evaluation on Synthetic Data}
For quantitative comparison, we trained \li\ and \zhu\ on our OpenRooms FF. Since \zhu's LightNet training code is not available, we used a pre-trained model trained on their InteriorVerse dataset for LightNet. Tab.~\ref{tab:compare} shows the performance comparison for all test images in OpenRooms FF. Although \li\ and \zhu\ are identical except for the network architecture, there is a significant difference in material and geometry performance between the two methods. This shows that \zhu' DenseNet is efficient for inverse rendering. Likewise, since the performance of MAIR++'s MGNet was superior to MAIR's NormalNet, MAIR also used MGNet's normal map for a fair comparison. Material, geometry, and rendering errors were measured by MSE and lighting error was measured by log space MSE with ground truth per-pixel environment map. Our method outperformed the previous method in all measurements, showing particularly overwhelming performance in re-rendering due to ILR's neural rendering. Because our ILR, like \li\ , obtains a 2D per-pixel lighting separately, we recorded it as $\mathrm{E}$-2D.
We provide inverse rendering results for OpenRooms FF test scene in Fig.~\ref{fig:ir_syn}. Although strong shadows appeared in the image bed, our method was able to predict the light quite successfully and remove it from the albedo. In addition, we reproduced the image almost as is. 

We also provide qualitative comparison results for per-pixel lighting in Fig.~\ref{fig:eval_light}. \zhu\ fails to predict the direction of the light source, and \li only predicts the approximate direction of the light source and does not accurately predict the surrounding lighting. In the case of ILR, it predicted the light source direction more accurately than existing methods but was unable to clearly express indirect lighting caused by surrounding objects. In the case of MAIR and MAIR++'s VSG, indirect lighting is expressed more realistically than per-pixel lighting because the entire scene is modeled with a 3d lighting volume.

\begin{table}[ht]
\footnotesize
\centering
\begin{tabular}{|l|c|c|c|c|}
\hline
MSE                          &\li    &\zhu    & MAIR  & MAIR++            \\ \hline
$\mathrm{N}$ $\downarrow$    & 3.782 & 1.738  & \multicolumn{2}{c|}{1.01} \\ \hline
$\mathrm{D}$ $\downarrow$    & 0.135 & 0.062  & 0.14  & \textbf{0.007} \\ \hline
$\mathrm{A}$ $\downarrow$    & 0.869 & 0.528  & 0.639 & \textbf{0.478} \\ \hline
$\mathrm{R}$ $\downarrow$    & 6.496 & 4.208  & 3.457 & \textbf{2.497} \\ \hline
$\mathrm{E}$-2D $\downarrow$ & 15.49 & -      & -     & \textbf{11.63} \\ \hline
$\mathrm{E}$-3D $\downarrow$ & -     & *31.15 & 13.79 & \textbf{11.61} \\ \hline
$\mathrm{I}$ $\downarrow$    & 0.491 & *3.565 & 0.784 & \textbf{0.058} \\ \hline
\end{tabular}
\caption{Quantitative comparison of material, geometry, and lighting in OpenRooms FF. The unit is $\times 10^{-2}$. * indicates that it was trained on the InteriorVerse dataset.}
\label{tab:compare}
\end{table}

\begin{figure*}[ht]
  \centering
  \includegraphics[width=\linewidth]{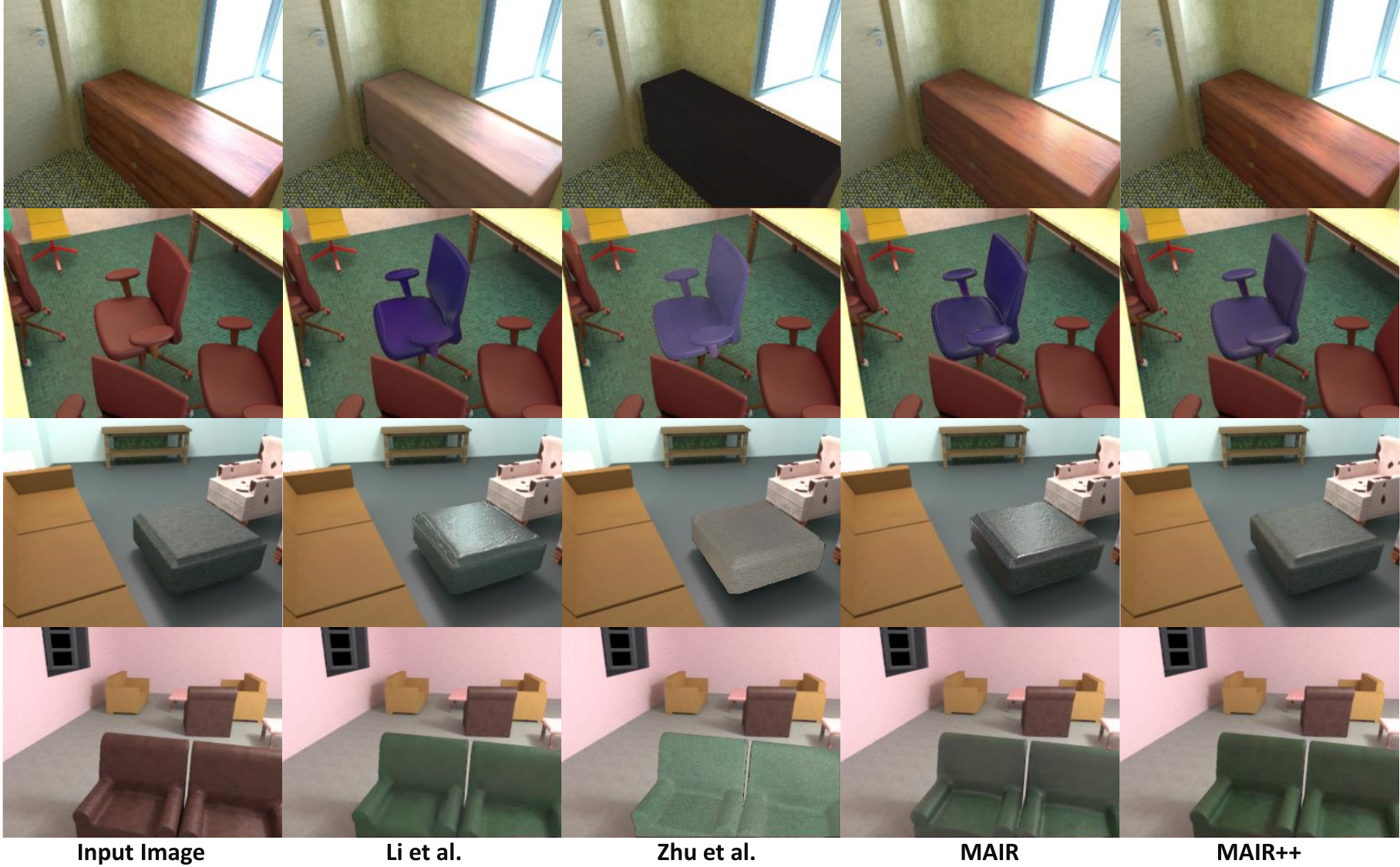}
   \caption{Material editing results for OpenRooms FF test data. The first row increases the roughness to weakens specularity, the second row increases the blue color of the diffuse albedo and strengthens specularity, the third row strengthens specularity, and the fourth row changes the diffuse albedo.} 
   \label{fig:mat_edit_syn}
\end{figure*}

\begin{figure*}[ht]
  \centering
  \includegraphics[width=\linewidth]{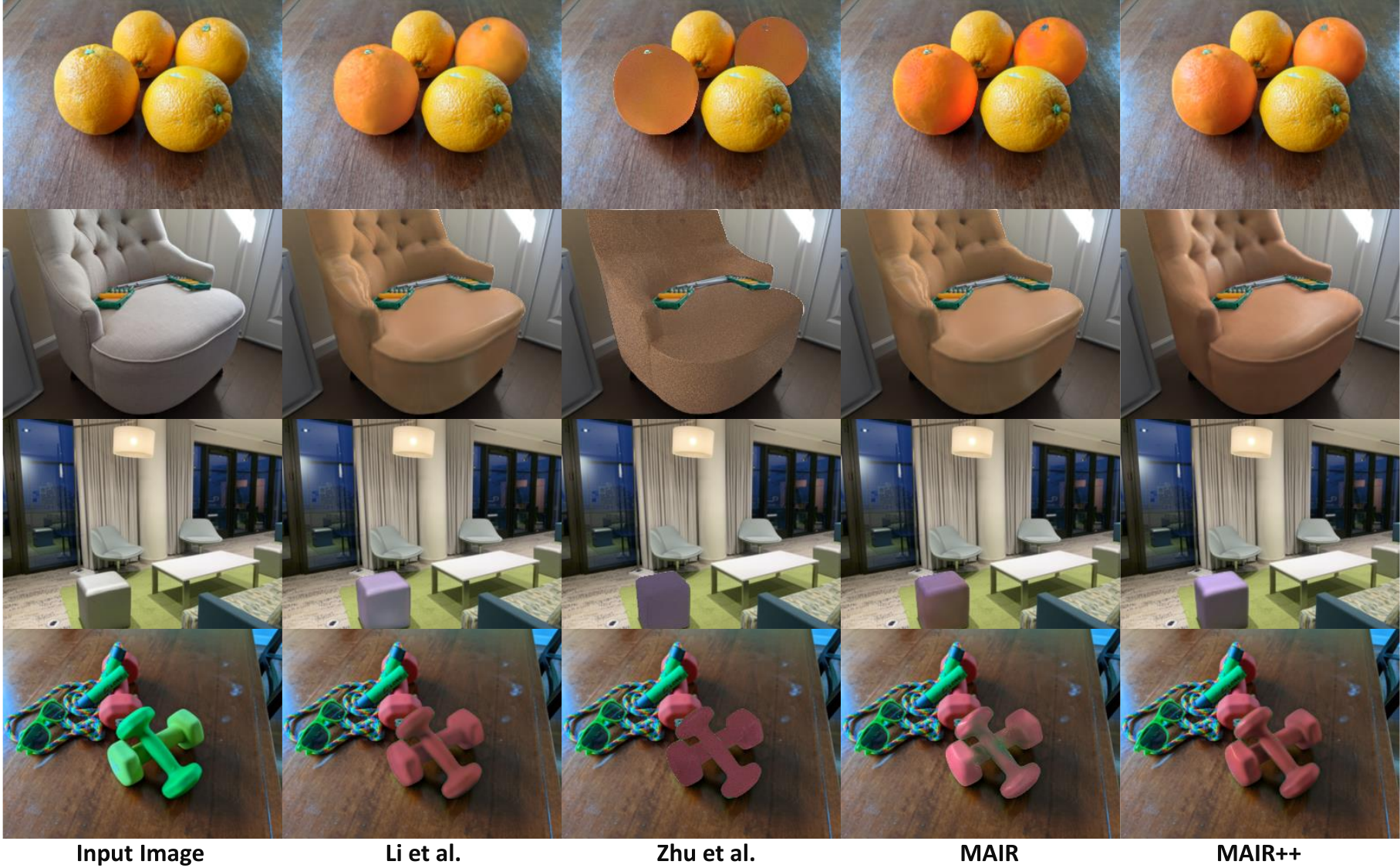}
   \caption{Material editing results for unseen real-world data. The first row changes the diffuse albedo, the second row changes the diffuse albedo and strengthens specularity, the third row changes the diffuse albedo and weakens specularity, and the fourth row changes the diffuse albedo.} 
   \label{fig:mat_edit_real}
\end{figure*}

\subsection{Evaluation on Real-world Data}
We evaluate the performance of inverse rendering for unseen real-world scenes in the IBRNet dataset\cite{ibrnet} in which the context of the scene is difficult to grasp. The performance gaps between multi-view methods and the single-view methods are more distinct in the unseen real-world scene. As shown in Fig.~\ref{fig:ir_real}, \li\ fails in geometry estimation due to the lack of context of the scene, leading to inaccurate disentanglement of the material and illumination. \zhu\ predicted geometry more accurately than \li\ thanks to its powerful DenseNet. However, they failed to infer lighting and as a result predicted the diffuse albedo to be too bright. On the other hand, MAIR and MAIR++ obtained fairly accurate geometry despite being unseen data thanks to MVS depth, which led to accurate lighting estimation. However, the specular component still remained in MAIR's albedo. MAIR++ was able to eliminate this more successfully by focusing more on the relationship between multi-views with ILR.

\begin{figure*}[ht]
  \centering
  \includegraphics[width=\linewidth]{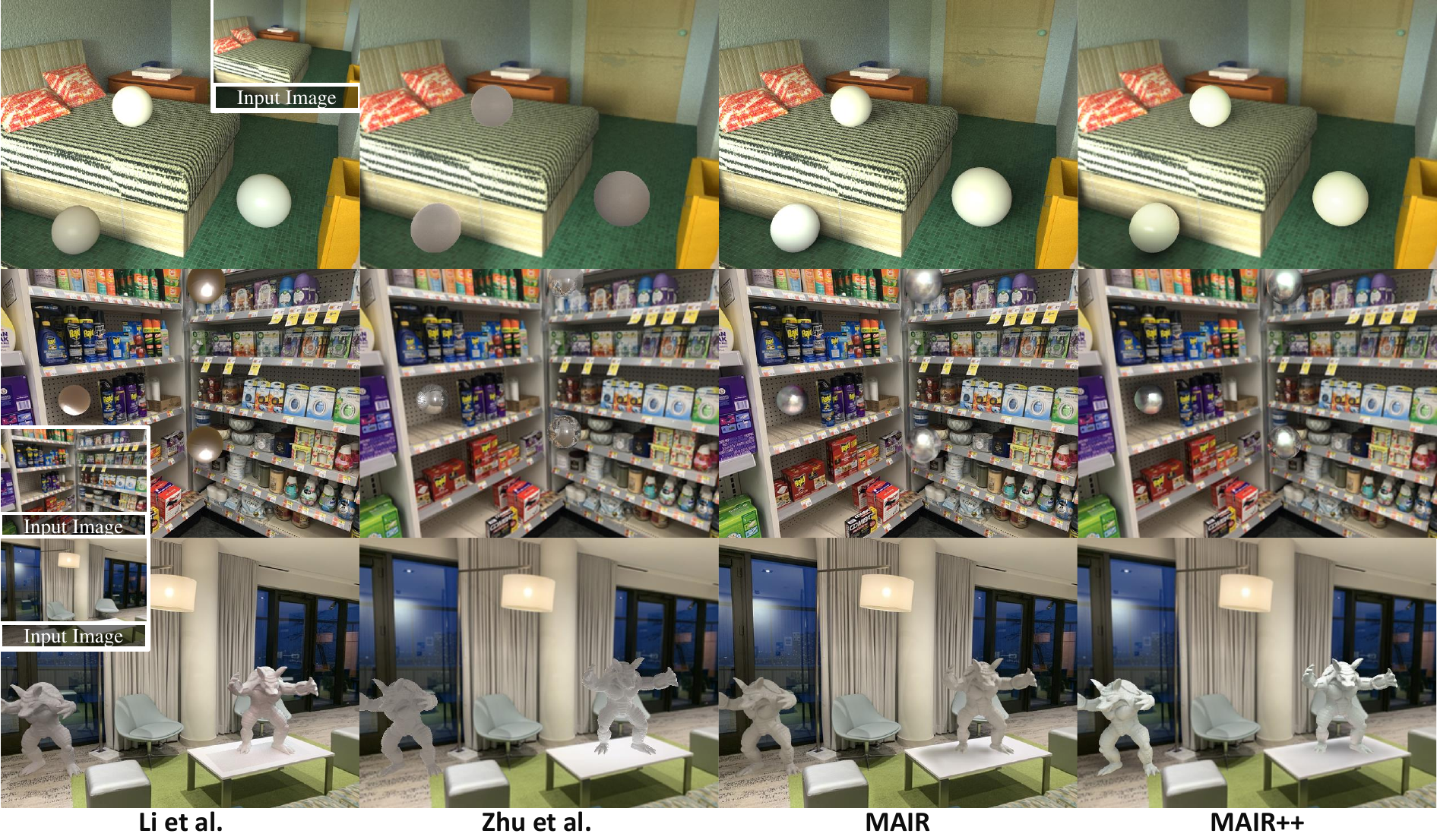}
  \caption{Object insertion experimental results. First row: Inserting a white sphere into the OpenRooms test scene. Second row: Inserting a chrome sphere into an unseen real world scene. Third row: Inserting a white armadillo~\cite{stanford} into an unseen real world scene.}
  \label{fig:oi_all}
\end{figure*}

\subsection{Evaluation on Material Editing}
Our ILR’s neural rendering enables realistic material editing. To verify this, we experimented with it in each dataset. To edit the material, we changed the material map obtained by each method and re-rendered it with each method's lighting. The diffuse albedo was adjusted by changing the scale, swapping channels, or adding offset, and the roughness was set to a value between 0 and 1 to control the specular component. \li\ and MAIR used uniform sampling rendering, \zhu\ used importance sampling provided by them, and MAIR++ used ILR neural rendering. We used SAM~\cite{kirillov2023segment} as the image segmentation model. 

Fig.~\ref{fig:mat_edit_syn} is the result of an experiment in the OpenRooms test scene. \zhu\ uses importance sampling, which is an improvement over uniform sampling rendering, but failed to properly infer lighting and showed inadequate results in all experiments. In the first row, we tried to remove the specular by setting the drawer's roughness to 1.0, and MAIR++ did this most successfully. In the second row, we changed the chair color to blue and enhanced the specular. Due to the limitations of uniform sampling rendering, \li\ and MAIR show artifacts at the edges of the chair and unnatural specular, while MAIR++ provides realistic rendering results. In the third row, we strengthen the specular. In \li\ and MAIR, the specular was overrepresented and artifacts were discovered. In the fourth row, we only changed the albedo of the chair. \li\ has lost the specular component of the original chair, while MAIR still had artifacts on the edges and made the chair brighter than before. 

This difference becomes more apparent in unseen real-world scenes, where inverse rendering is difficult. In the first row of Fig.~\ref{fig:mat_edit_real} we have changed the albedo of the two oranges. While \li\ and MAIR lose the specular component of the original orange, MAIR++ successfully reproduces it and shows realistic rendering. In the second row, we changed the albedo of the chair to be more red and added a specular component. \li\ and MAIR show artifacts in the handle of the chair. MAIR++ reproduces the chair's shadow most realistically. In the third row, we changed the albedo of the chair to purple and removed the specular. Although \li and MAIR did not reproduce the shading of the chair properly, MAIR++ not only removed the specular but also expressed the strong shading of the chair. In the fourth row, we changed the albedo of the dumbbell. \li\ could not properly express the existing shading, and MAIR extracted the lighting from VSG, so the green light of the existing indirect lighting was revealed. MAIR++ edits materials in the most realistic way.

\subsection{Evaluation on Object Insertion}
We further demonstrate the effectiveness of our robust inverse rendering in the object insertion task. In Fig.~\ref{fig:oi_all} we provide a comparison with the single-view methods. We implemented a simple renderer for object insertion by referring to Wang \etal\cite{vsg} and used it to render the results of MAIR and MAIR++. As the public implementation of \li\ includes a renderer of its own, the results of \li were rendered using this renderer, except for the results of the chrome sphere insertion; the renderer from \li\ does not support the rendering of the chrome sphere directly, so we used our renderer for this case. In \li's object insertion application, the user must specify the plane on which the object is located. Therefore, an object cannot be placed on another object or floated in the air, and shadows are only cast on the plane. In our method, it is possible to insert objects freely into the 3D space without restrictions on the shadow. We applied the VSG obtained for the center-view to object insertion. \zhu's importance sampling renderer supports all object insertions, including chrome spheres, so the results of \zhu were created using this renderer. However, this renderer did not support shadows. 

The first row of Fig.~\ref{fig:oi_all} shows the result of inserting a white sphere into the OpenRooms test scene. \zhu\ predicted the lighting too dark and showed unnatural results, and \li\ rendered the sphere on the left side of the floor too dark and could not even express the shadow of the sphere. MAIR and MAIR++ successfully predicted 3D lighting, appropriately expressing the brightness of the sphere and also expressing shadows. In the second row, we have inserted a chrome sphere into the real world scene. \li\ gave unnatural results for this and the sphere of \zhu showed that the surrounding lighting was not taken into account. Ambient lighting is more clearly visible on MAIR's chrome sphere, but the spheres in the darker areas within the shelf have an awkward purple glow reflecting off them. On the other hand, MAIR++'s chrome sphere reflects light more naturally. The third row shows the result of inserting the Stanford Armadillo~\cite{stanford} into a real-world scene. \li's method did not properly express the shadow of the object, and the method of \zhu inserted the object in an unrealistic way. MAIR inserted objects too darkly without considering indoor lighting. On the other hand, MAIR++ realistically expressed the shading of objects.

\subsection{Ablation Study}
\noindent{\bf Design of stage 2.} Tab.~\ref{tab:abl_BRDF_MAIR} shows experimental results for network design choices in stage 2. It shows that MVANet requires a local context of ContextNet for material estimation. In addition, RefineNet is necessary to compensate for the weakness of the pixel-wise operation in MVANet. We also validate the effect of $\boldsymbol{f}_{\text{spec}}$.``w/o $\boldsymbol{f}_{\text{spec}}$'' is the result of the model trained without $\boldsymbol{f}_{\text{spec}}$, and ``w/o reparameterize'' is the result of the model trained without physically motivated encoding (Eq.~\ref{eqn:fspec}) as follows:

\begin{equation}\label{eqn:fspec_abl}
\boldsymbol{f}_{\text{spec}}^{k} = \displaystyle\sum_{s=1}^{S_D} \text{SpecNet}(\boldsymbol{v}_k, \Tilde{\boldsymbol{n}}, \boldsymbol{\xi}_s, \boldsymbol{\eta}_s, \lambda_s).
\end{equation}

Improvement in roughness estimation implies that our physically motivated encoding considering the microfacet BRDF~\cite{microfacet} model helps in estimating specular radiance. Additionally, we confirmed through "MAIR with $\mathrm{D}_t$" that accurate depth is helpful for multi-view material estimation.

\begin{table}[htb]
\centering
\begin{tabular}{|l|c|c|}
\hline
Components                            & albedo($\mathrm{A}$)$\downarrow$    & roughness($\mathrm{R}$)$\downarrow$      \\ \hline
MAIR                                  & \underline{0.639}  & \underline{3.457} \\ \hline
MAIR with $\mathrm{D}_t$              & \textbf{0.621}  & \textbf{3.344}          \\ \hline
w/o ContextNet                        & 0.726 & 5.409          \\ \hline
w/o RefineNet                         & 0.94  & 6.059          \\ \hline
w/o $\boldsymbol{f}_\text{spec}$      & 0.713 & 4.391          \\ \hline
w/o reparameterize                    & 0.665 & 4.457          \\ \hline
\end{tabular}
\caption{Ablation studies in stage 2 of MAIR. The best results are in bold, and the second best results are underlined.}
\label{tab:abl_BRDF_MAIR}
\end{table}

\begin{figure*}[ht]
  \centering
  \includegraphics[width=\linewidth]{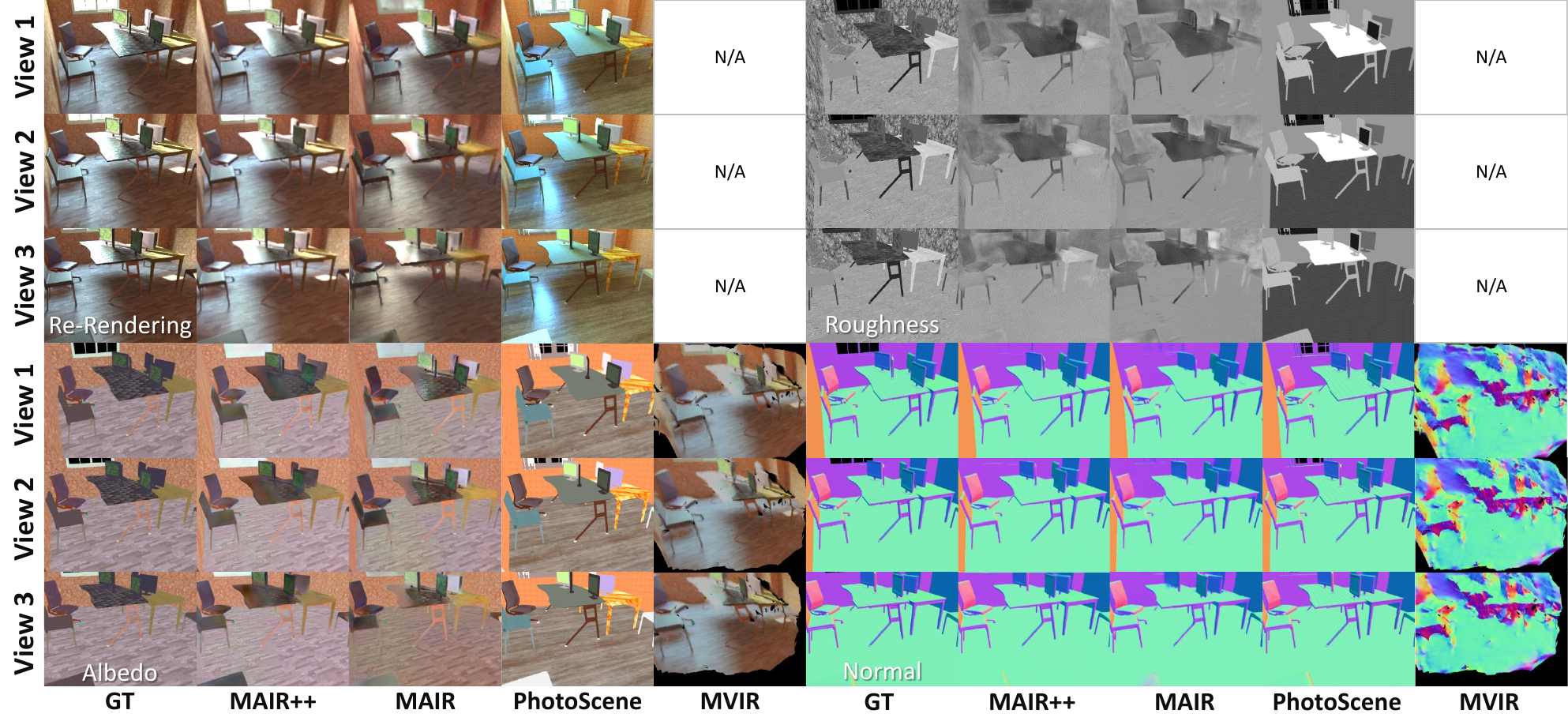}
  \caption{Comparisons with PhotoScene~\cite{photoscene} and MVIR~\cite{kim2016multi} in OpenRooms FF test scene.}
  \label{fig:comparisons}
\end{figure*}

\noindent{\bf Performance of ILRNet.} Tab.~\ref{tab:abl_GT_ILR} shows that it is appropriate to use ground truth to prevent overfitting to the input image when training ILRNet's neural renderer. "w/o GT" is the result of learning using MGNet's material and geometry without using ground truth. "w/o GT" has lower RGB rendering loss while higher diffuse rendering loss. This means that the shading map of ILRNet's integrator contains part of the input image. Additionally, “MVM+DAM(with no GT ILR)” in Tab.~\ref{tab:abl_BRDF_MAIR++} shows that ILRNet overfitted to the input image has a negative effect on material estimation. 

We also quantitatively evaluated the accuracy and speed of ILRNet's neural rendering. Tab.~\ref{tab:rendering} shows the experimental results for both datasets. USR($\widetilde{\mathrm{A}}, \widetilde{\mathrm{R}}, \widetilde{\mathrm{N}}, \widetilde{\mathrm{E}}$) means uniform sampling rendering using ground truth material, geometry, and per-pixel lighting. USR inevitably produces artifacts due to the limitations of rendering itself (especially when directional resolution is low), and as a result, it shows more inaccurate results than ILRNet. Additionally, ILRNet shows accurate rendering performance and fast rendering speed ahead of existing methods. 

\begin{table}[htb]
\centering
\begin{tabular}{|l|c|c|c|c|}
\hline
Method                & $\mathrm{I}_d\downarrow$    & $\mathrm{I}_s\downarrow$ & $\mathrm{I}\downarrow$ & $\mathrm{E}\downarrow$      \\ \hline
ILRNet                    & 0.321                                  & 0.160 & 0.347 & 11.63 \\ \hline
ILRNet (w/o GT)           & 0.539                                  & 0.157 & 0.176 & 11.78 \\ \hline
\end{tabular}
\caption{Ablation study on the necessity of using ground truth material and geometry when training ILRNet's neural renderer. The unit of loss is si-MSE ($\times 10^{-2}$).}
\label{tab:abl_GT_ILR}
\end{table}

\begin{table}[htb]
\centering
\renewcommand{\arraystretch}{1.5} 
\begin{tabular}{|l|c|c|c|}
\hline
& OpenRooms FF & IBRNet & Runtime \\ \hline
\li         & 0.491          & \underline{1.145}          & 100ms  \\ \hline
\zhu        & 3.565          & 3.517          & 12s    \\ \hline
MAIR        & 0.784          & 1.173          & 792ms      \\ \hline
MAIR++      & \textbf{0.058} & \textbf{0.321} & \underline{81ms}   \\ \hline
USR($\widetilde{\mathrm{A}}, \widetilde{\mathrm{R}}, \widetilde{\mathrm{N}}, \widetilde{\mathrm{E}}$) & \underline{0.274} & - & \textbf{2ms} \\ \hline
\end{tabular}
\caption{Re-rendering accuracy (MSE $\times 10^{-2}$) and processing time.}
\label{tab:rendering}
\end{table}

\noindent{\bf Design of MVANet.} Together with ILR and the DAM, we were able to infer material more accurately. To verify its performance, we thoroughly evaluated MVANet of various case designs in Tab.~\ref{tab:abl_BRDF_MAIR++}. MVM means Mean-Variance Module, and MVM+DAM means using both modules in parallel. $\boldsymbol{w}$ is the multi-view weight in Eq.~\ref{eqn:multi_view_weight} and $\boldsymbol{m}$ is the multi-view mask in Eq.~\ref{eqn:multi_view_mask}. Unless otherwise specified, $\boldsymbol{w}$ is used for MVM and $\boldsymbol{m}$ is used for DAM. “w/o $\boldsymbol{t}$” means that DAM uses the token of the target view as output instead of target embedding $\boldsymbol{t}$, which slightly reduces performance. In the DAM, using masked attention("MVM+DAM") and weighted attention("MVM+DAM(with $\boldsymbol{w}$)") showed similar performance, but we chose masked attention because it is slightly more computationally efficient. "w/o PE" means concatenating $\boldsymbol{f}_{\text{spec}}^{k}$, $\mathrm{I}^k$, and $\boldsymbol{f}_{\text{context}}$ without adding $\boldsymbol{f}_{\text{spec}}^{k}$ to $\mathrm{I}^k$ and $\boldsymbol{f}_{\text{context}}$ like in Eq.~\ref{eqn:da_pe}. PE made a significant contribution to inferring the roughness map. We also tested whether SpecRenderer features and $\mathrm{S}$ could replace SpecNet. The results of "with SpecRenderer" show that SpecRenderer's features are not very helpful. We speculated that this is because the information for rendering and the light information needed to infer the material are different. Additionally, the results of MVM(w/o $\boldsymbol{w}$) and DAM(w/o $\boldsymbol{m}$) imply that it is very important to consider occlusion in multi-view material estimation. We also experimented with RNN (Recurrent Neural Network)-based structures, which showed good performance in processing time series data. However, it did not show as much performance as MVM or DAM. Finally, we verified whether ILR is more helpful for material estimation than SVSGs. SVSGs used 12 SGs for each pixel, and were implemented with DenseNet, the same as ILRNet's LightEncoder. In all three experiments (MVM, DAM, and MVM+DAM), ILR demonstrated a significant performance improvement compared to SVSGs, indicating that ILR provides richer information about light than SVSGs.

\begin{table}[htb]
\centering
\begin{tabular}{|cl|l|l|l|}
\hline
\multicolumn{2}{|c|}{Components}                & \multicolumn{1}{c|}{$\mathrm{A}\downarrow$}         & \multicolumn{1}{c|}{$\mathrm{R}\downarrow$}         & \multicolumn{1}{c|}{$\mathrm{I}\downarrow$} \\ \hline

\multicolumn{1}{|c|}{\multirow{10}{*}{ILR}}   & MVM+DAM  &  \textbf{0.507} & \multicolumn{1}{c|}{\underline{2.497}} & \multicolumn{1}{c|}{\textbf{0.038}}   \\ \cline{2-5} 
\multicolumn{1}{|c|}{}                        & MVM+DAM(w/o $\boldsymbol{t}$)  & \underline{0.510} & 2.658 & \underline{0.040}  \\ \cline{2-5} 
\multicolumn{1}{|c|}{}                        & MVM+DAM(with $\boldsymbol{w}$) & 0.513   & \textbf{2.469}  & \underline{0.040}  \\ \cline{2-5} 
\multicolumn{1}{|c|}{}                        & MVM+DAM(with $\boldsymbol{w}$, w/o $\boldsymbol{t}$) & \underline{0.510} & 2.571  & \textbf{0.038} \\ \cline{2-5} 
\multicolumn{1}{|c|}{}                        & MVM+DAM(w/o PE)                & 0.513   & 2.900           & 0.041           \\ \cline{2-5} 
\multicolumn{1}{|c|}{}                        & MVM+DAM(with SpecRenderer)     & 0.523   & 3.184           & 0.097            \\ \cline{2-5} 
\multicolumn{1}{|c|}{}                        & MVM+DAM(with no GT ILR)        & 0.513   & 2.697           & 0.056            \\ \cline{2-5} 
\multicolumn{1}{|c|}{}                        & MVM                           & 0.525   & 2.919           & 0.041           \\ \cline{2-5} 
\multicolumn{1}{|c|}{}                        & MVM(w/o $\boldsymbol{w}$)     & 0.528   & 3.972           & 0.044          \\ \cline{2-5} 
\multicolumn{1}{|c|}{}                        & DAM                           & 0.512   & 2.748           & 0.041           \\ \cline{2-5} 
\multicolumn{1}{|c|}{}                        & DAM(w/o $\boldsymbol{m}$)     & 0.531   & 5.754           & 0.046          \\ \cline{2-5} 
\multicolumn{1}{|c|}{}                        & RNN                          & 0.538   & 6.753           & 0.494           \\ \hline
\multicolumn{1}{|c|}{\multirow{3}{*}{SVSGs}}  & MVM+DAM                        & 0.511   & 2.71            & 0.255            \\ \cline{2-5} 
\multicolumn{1}{|c|}{}                        & MVM                           & 0.547   & 3.236           & 0.275            \\ \cline{2-5} 
\multicolumn{1}{|c|}{}                        & DAM                           & 0.523   & 3.179           & 0.258            \\ \hline
\end{tabular}
\caption{Ablation studies in stage 2 of MAIR++.}
\label{tab:abl_BRDF_MAIR++}
\end{table}


\section{Discussion}
\noindent{\bf Comparisons with 3D geometry-based methods.}
Fig.~\ref{fig:comparisons} shows the comparison results with PhotoScene~\cite{photoscene} and MVIR~\cite{kim2016multi} in 3 views. The inference time is 2s for MAIR and MAIR++, 9m 52s for MVIR~\cite{kim2016multi} and 10m 40s for PhotoScene~\cite{photoscene} on RTX 2080 Ti. MVIR~\cite{kim2016multi} fails to generate geometry, showing severe artifacts. The PhotoScene~\cite{photoscene} shows a complete scene reconstruction owing to the CAD geometry and material graph, but the synthesized images largely differ from the original scene. 

\noindent{\bf Inter-view consistency.}
Unlike 3D geometry-based methods~\cite{photoscene, kim2016multi}, which leverage global 3D geometry to achieve consistency, MAIR and MAIR++ operate in pixel-space and therefore does not explicitly guarantee inter-view consistency. However, we did not observe significant inconsistencies during our experiment, presumably due to the role of MVANet, which can be seen in Fig.~\ref{fig:comparisons}. 

\noindent{\bf 3D lighting volume with ILR.} Although we have successfully expressed 2D surface lighting with ILR, 3D volume lighting still relies on the existing VSG representation. We expect that expressing 3D volume lighting with ILR will show more realistic results in object insertion applications. However, simply replacing VSG with ILR has several problems, such as requiring a lot of resources, so we leave this as future work.

\begin{figure}[ht]
  \centering
  \includegraphics[width=\linewidth]{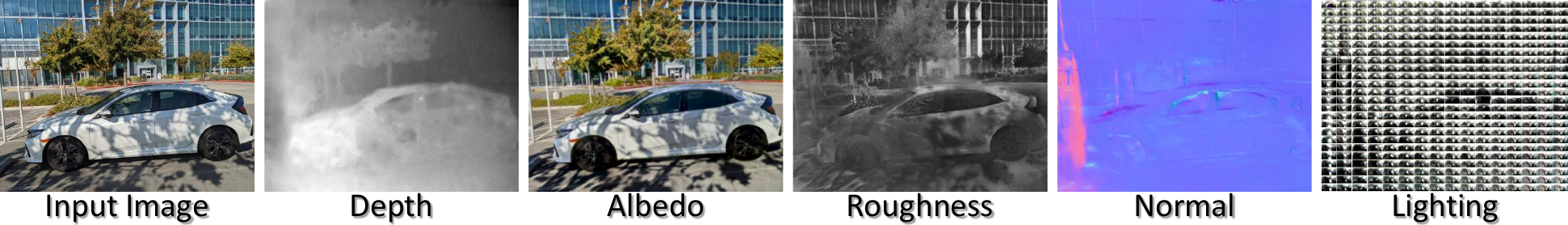}
  \caption{Failure case when depth prediction fails.}
  \label{fig:failure_case}
\end{figure}

\noindent{\bf Limitation.}
One possible limitation comes from the cascaded nature of the pipeline. If depth estimation fails due to, for example, the presence of dynamic objects or large textureless region, MAIR and MAIR++ will not work properly (see Fig.~\ref{fig:failure_case}.). Another possible limitation comes from the VSG representation. Although VSG can express 3D lighting effectively, it cannot be applied to applications such as light source editing because it is non-parametric. 

\section{Conclusion}
We further expanded the application range of MAIR based on implicit lighting representation. Compared to existing methods, we were able to more accurately decompose images into material, geometry, and spatially-varying lighting, which could be successfully applied to a variety of applications. In particular, ILR's neural rendering greatly improved re-rendering performance, making realistic material editing possible. Our method showed stable performance even in real-world scenes, which could be practically applied to VR / AR fields.


\noindent\textbf{Acknowledgements.}
This work was supported by Institute of Information \& communications Technology Planning \& Evaluation (IITP) grant 
funded by the Korea government(MSIT)(No.RS-2023-00227592, Development of 3D Object Identification Technology Robust to Viewpoint Changes)

\bibliographystyle{IEEEtran}
\bibliography{egbib}

\end{document}